%% file: main.tex
\def\isarxiv{1} 

\ifdefined\isarxiv
\documentclass[11pt]{article}

\usepackage[numbers]{natbib}

\else

\documentclass[twoside]{article}

\usepackage[accepted]{aistats2025}
\usepackage[round]{natbib}

\fi

\usepackage{amsmath}
\usepackage{amsthm}
\usepackage{amssymb} 
\usepackage{algorithm}
\usepackage{algpseudocode}
\usepackage{grffile}
\usepackage{wrapfig,epsfig}
\usepackage{url}
\usepackage{xcolor}
\usepackage{color}
\usepackage{epstopdf}
\usepackage{bbm}
\usepackage{dsfont}
\allowdisplaybreaks

\ifdefined\isarxiv

\usepackage{tikz}
\usepackage{hyperref}  
\hypersetup{colorlinks=true,citecolor=blue,linkcolor=blue} 
\usetikzlibrary{arrows}
\usepackage[margin=1in]{geometry}

\else

 \usepackage{microtype}
\usepackage{hyperref}
 \definecolor{mydarkblue}{rgb}{0,0.08,0.45}
 \hypersetup{colorlinks=true, citecolor=mydarkblue,linkcolor=mydarkblue}

\fi

\theoremstyle{plain}
\newtheorem{theorem}{Theorem}[section]
\newtheorem{lemma}[theorem]{Lemma}
\newtheorem{definition}[theorem]{Definition}

\newtheorem{corollary}[theorem]{Corollary}

\newtheorem{fact}[theorem]{Fact}

\newtheorem{claim}[theorem]{Claim}

\newcommand{\wh}{\widehat}
\newcommand{\wt}{\widetilde}
\newcommand{\ov}{\overline}

\newcommand{\R}{\mathbb{R}}

\renewcommand{\d}{\mathrm{d}}

\DeclareMathOperator*{\Z}{\mathbb{Z}}

\DeclareMathOperator{\poly}{poly}

\DeclareMathOperator{\nnz}{nnz}

\DeclareMathOperator{\rank}{rank}
\DeclareMathOperator{\diag}{diag}

\DeclareMathOperator{\reg}{reg}

\makeatletter
\newcommand*{\RN}[1]{\expandafter\@slowromancap\romannumeral #1@}
\makeatother

\usepackage{lineno}

\ifdefined\isarxiv

\title{An Iterative Algorithm for Rescaled Hyperbolic Functions Regression}

\else

\fi

\begin{document}

\ifdefined\isarxiv

\date{}

\author{
Yeqi Gao\thanks{\texttt{a916755226@gmail.com}. The University of Washington.}
\and 
Zhao Song\thanks{\texttt{magic.linuxkde@gmail.com}. Simons Institute for the Theory of Computing, UC Berkeley.}
\and
Junze Yin\thanks{\texttt{jy158@rice.edu}. Rice University.}
}

\else

\twocolumn[

\aistatstitle{An Iterative Algorithm for Rescaled Hyperbolic Functions Regression}

\aistatsauthor{ Yeqi Gao \And Zhao Song \And  Junze Yin }

\aistatsaddress{ The University of Washington \And  Simons Institute for the Theory\\ of Computing, UC Berkeley \And Rice University } ]

\fi

\ifdefined\isarxiv
\begin{titlepage}
  \maketitle
  \begin{abstract}
\input{abstract}

  \end{abstract}
  \thispagestyle{empty}
\end{titlepage}

{\hypersetup{linkcolor=black}
}
\newpage

\else
\begin{abstract}
\input{abstract}
\end{abstract}

\fi

\input{intro} 
\input{related}

\input{short_preli}
\input{tech}

\input{short_hessian}
\input{short_newton}

\input{conclusion}

\ifdefined\isarxiv

\else
\bibliographystyle{plainnat}
\bibliography{ref}
\input{checklist}
\fi

\onecolumn
\appendix

\ifdefined\isarxiv

\else
\aistatstitle{An Iterative Algorithm for Rescaled Hyperbolic Functions Regression:\\
Supplementary Materials}
\fi

\input{preli}
\input{g_H}
\input{pd}
\input{lip}

\input{lip_A}
\input{result}



\ifdefined\isarxiv
\bibliographystyle{alpha}
\bibliography{ref}
\else

\fi



\end{document}

%% file: abstract.tex
Large language models (LLMs) have numerous real-life applications across various domains, such as natural language translation, sentiment analysis, language modeling, chatbots and conversational agents, creative writing, text classification, summarization, and generation. LLMs have shown great promise in improving the accuracy and efficiency of these tasks, and have the potential to revolutionize the field of natural language processing (NLP) in the years to come.
Exponential function based attention unit is a fundamental element in LLMs. Several previous works have studied the convergence of exponential regression and softmax regression.

In this paper, we propose an iterative algorithm to solve a rescaled version of the slightly different formulation of the softmax regression problem that arises in attention mechanisms of large language models. Specifically, we consider minimizing the squared loss between a certain function, which can be either the exponential function, hyperbolic sine function, or hyperbolic cosine function, and its inner product with a target $n$-dimensional vector $b$, scaled by the normalization term. This ``rescaled softmax regression'' differs from classical softmax regression in the location of the normalization factor.

The efficiency and generalizability of this framework to multiple hyperbolic functions make it relevant for optimizing attention mechanisms. The analysis also leads to a corollary bounding solution changes under small perturbations for in-context learning. Limitations and societal impact are discussed.

%% file: intro.tex
\section{Introduction}
The background of large language models (LLMs) can be traced back to a series of groundbreaking models, including the Transformer model \citep{vsp+17}, GPT-1 \citep{rns+18}, BERT \citep{dclt18}, GPT-2 \citep{rwc+19}, and GPT-3 \citep{bmr+20}. These models are trained on massive amounts of textual data to generate natural language text and have shown their power on various real-world tasks, including natural language translation \citep{hwl21}, sentiment analysis \citep{uas+20}, language modeling \citep{mms+19}, and even creative writing \citep{o23}. The success of the new version of LLM named GPT-4 \citep{o23} has exemplified the use of LLMs in human-interaction tasks and suggests that LLMs are likely to continue to be a key area of research in the years to come.

LLMs rely heavily on attention computations to improve their performance in natural language processing tasks. The attention mechanism enables the model to selectively focus on specific parts of the input text \citep{vsp+17, dclt18, rwc+19, bmr+20, rns+18}, enhancing its ability to identify and extract relevant information. A crucial component of the attention mechanism is the attention matrix, a square matrix in which each entry represents the correlations between words or tokens in the input text. The entries in the matrix are computed using a soft attention mechanism, which generates weights by applying a softmax function over the input sequence. Through this process, LLMs can identify and prioritize important parts of the input text, resulting in more accurate and efficient language processing.

Mathematically, one layer of forward computation is defined as follows:
\begin{definition}[$\ell$-th layer forward computation and attention optimization]\label{def:attention_comp}
Let $n$ denote the length of the input token, and $d$ denote the hidden dimension.

For ${\bf 1}_n$ being a vector whose entries are all $1$'s and dimension is $n$, $\diag$ being a function mapping a vector in $\R^n$ to a matrix in $\R^{n \times n}$ (each of the entries of the vector in $\R^n$ is mapped to the diagonal entries of the $n \times n$ matrix), $Q, K, V \in \R^{d \times d}$ being the weights of the query, key, and value, respectively, $X_{\ell} \in \R^{n \times d}$ being the $\ell$-th layer input, the $\ell$-th layer forward computation is
\begin{align*}
 X_{\ell+1} \gets D^{-1} \exp(X_{\ell} Q K^\top X_{\ell}^\top) X_{\ell} V,
\end{align*} 
where $D:= \diag( \exp( X_{\ell} Q K^\top X_{\ell}^\top ){\bf 1}_n )$.
\end{definition}

Therefore, the attention optimization is defined as
\begin{align}\label{eq:attention}
    \min_{X,Y \in \R^{d \times d}}  \| D(X)^{-1} \exp(A_1 X A_2^\top) A_3 Y - B \|_F^2,
\end{align}
where $\|\cdot\|_F$ is the Frobenius norm, $QK^\top$ is merged into $X \in \R^{d \times d}$, and $Y = V \in \R^{d \times d}$ are the weights which are interested to learn. $D(X) = \diag( \exp(A_1 X A_2^\top ) {\bf 1}_n ) \in \R^{n \times n}$ and $A_1,A_2,A_3,B \in \R^{n \times d}$.

The attention mechanism has a computational complexity of $\wt{O}(n^2)$ with respect to the input sequence length $n$. The quadratic complexity of the attention computation makes it challenging for LLMs to efficiently process very long input sequences, which further limits the efficiency of training LLMs. Consequently, there has been growing interest in addressing the quadratic computational complexity by analyzing various regression problems derived from the attention computation (Definition~\ref{def:attention_comp}). Several recent studies investigate the computation of the attention matrix in LLMs, including \cite{zhdk23,bsz23,dms23,as23}. Specifically, \cite{zhdk23,bsz23,as23} explore:
\begin{align*}
    D^{-1} \exp(Q K^\top) V,
\end{align*}
where compared to Eq.~\eqref{eq:attention}, $A_1 X$ is merged into one matrix $Q$ and $A_3 Y$ is merged into one matrix $V$. To get an almost linear time algorithm to approximate the attention optimization problem, \cite{as23} relies on strict assumptions that $d = O(\log n)$ and all entries of $Q, K, V$ are bounded by $o(\sqrt{\log n})$.

\cite{dms23}, on the other hand, studies
\begin{align*}
    D^{-1} \exp(A_2 A_2^\top),
\end{align*}
where $A_3Y$ is not considered and only considers the symmetric matrix. \cite{kmz23} also replaces the softmax function $\exp$ in the attention mechanism with polynomials. While simplifying the attention optimization problem is acceptable and can reduce quadratic complexity to accelerate the training of LLMs, making too many modifications will inevitably have a negative impact on their performance \citep{dlz+23}. Thus, there is a trade-off between the efficiency of LLM training and its performance. It is natural to ask:
\begin{center}
    {\it Is it possible to address quadratic computational complexity and accommodate more than the softmax unit with minimum simplifications to the attention optimization problem?}
\end{center}

In this work, we provide a positive answer to this question: we focus on and develop the direction of regression tasks from \cite{dls23,lsx+23}, called the softmax regression 
\begin{align*}
    \min_{x \in \R^d} \| \langle \exp(Ax) , {\bf 1}_n \rangle^{-1} \exp(Ax) - b \|_2,
\end{align*}
to define and analyze the following novel regression problem:
\begin{definition}[Rescaled Softmax Regression]\label{def:rescaled_softmax_regression}

Let $A \in \R^{n \times d}$ and $x \in \R^d$. Let $u(x)$ be applied entry-wise to the vector $x$ and $u(x) \in \{\exp(Ax),\cosh(Ax),\sinh(Ax)\}$.
Let $b \in \R^n$. The goal of the rescaled softmax regression problem is to solve
\begin{align*}
\min_{x \in \R^d} \| u(x) - \langle u(x) , {\bf 1}_n \rangle \cdot b \|_2,
\end{align*}
where ${\bf 1}_n$ is the $n$-dimensional vector whose entries are all $1$.
\end{definition}

Compared to \cite{dms23,zhdk23,bsz23}, the softmax regression from \cite{dls23} is the problem with the smallest change from the original attention optimization problem, where only $A_3 Y$ is not considered. For \cite{dms23}, $A_3 Y$ is not considered and only considered for the symmetric matrix, and for \cite{zhdk23,bsz23}, $A_1 X$ is merged into one matrix $Q$ and $A_3 Y$ is merged into one matrix $V$. We minimize the simplifications of the attention computation (Definition~\ref{def:attention_comp}) and design a sub-quadratic algorithm (Algorithm~\ref{alg:main:informal}), which may lead to faster training in transformer models with minimum sacrifice in their performance.

\paragraph{Our contributions.}

Our contributions can be summarized as follows:
\begin{itemize}
    \item The first contribution of this paper is defining and analyzing the rescaled version of the softmax regression problem (Definition~\ref{def:rescaled_softmax_regression}) and creating a randomized algorithm to solve it in subquadratic time of $n$ (Theorem~\ref{thm:main_formal} and Theorem~\ref{thm:main_informal}).
    \item We remark the major difference between classical softmax regression and our new rescaled softmax regression (Definition~\ref{def:rescaled_softmax_regression}) is the location of the normalization factor $\langle u(x), {\bf 1}_n\rangle$. Due to the difference, the analysis for rescaled softmax regression will be quite different. This is the second contribution of this work. 
    \item The third contribution of this paper is that our framework is very general and can handle several hyperbolic functions at the same time, including $\exp$, $\cosh$, and $\sinh$, which is comparable to \cite{kmz23} that handles polynomial function.
\end{itemize}

\subsection{Our Results}

Note that we follow the assumption that $\| b \|_2 \leq R$ as in \cite{lsz23}. The reason why \cite{dls23} can assume that $\| b \|_2 \leq 1$ is because they solve a normalized version. Therefore, in our re-scaled version, we only assume $\| b \|_2 \leq R$. Moreover, inspired by the empirical success of using weight decay in training transformers as explained in \cite{llr23}, we explore a regularized version of Definition~\ref{def:rescaled_softmax_regression}, namely
\begin{align}\label{eq:reg_problem}
     \min_{x \in \R^d} 0.5 \cdot \| u(x) - \langle  u(x) , {\bf 1}_n \rangle \cdot b \|_2^2 + 0.5 \cdot \| \diag(w) A x\|_2^2,
\end{align}
where ${\bf 1}_n, u(x), b \in \R^n$, $x \in \R^d$, and $A \in \R^{n \times d}$ are defined as in Definition~\ref{def:rescaled_softmax_regression}. Also, $w \in \R^n$ and $\diag(w) \in \R^{n \times n}$ is a diagonal matrix that moves the entries of $w$ to the diagonal entries of $\diag(w)$.

The informal version of our main result is presented as follows:
\begin{theorem}[Main Result, Informal version of Theorem~\ref{thm:main_formal}]\label{thm:main_informal}

Let $\epsilon, \delta \in (0, 0.1)$ be the accuracy parameter and the failure probability, respectively. 

Let $x_0, x^* \in \R^d$ denote the initial point and the optimal solution respectively, $\nnz(A)$ denote the number of non-zero entries of $A$, and $\omega\approx 2.37$. 

Then, there exists a randomized algorithm (Algorithm~\ref{alg:main:informal}) solving Eq.~\eqref{eq:reg_problem} such that, with probability at least $1-\delta$, runs $T = \log(\| x_0 - x^* \|_2/ \epsilon)$ iterations, spends 
\begin{align*}
    O( (\nnz(A) + d^{\omega} ) \cdot \poly(\log(n/\delta))
\end{align*}
time in each iteration, and outputs a vector $\wt{x} \in \R^d$ such that 
\begin{align*}
    \| \wt{x} - x^* \|_2 \leq \epsilon.
\end{align*}

\end{theorem}

\paragraph{Roadmap.}
Our paper is organized as follows. In Section~\ref{sec:related_work}, we discuss related work. In Section~\ref{sec:short_preli}, we introduce several basic mathematical notations that we use in this paper. In Section~\ref{sec:technique_overview}, we provide a technique overview. 
In Section~\ref{sec:short_hessian}, we present several properties of Hessian of loss functions. 
In Section~\ref{sec:newton:short}, we present an analysis of our regression algorithm.
In Section~\ref{sec:conclusion}, we provide a conclusion.

%% file: related.tex
\section{Related Work}
\label{sec:related_work}

\paragraph{Optimization and Convergence}
Studies in the field of optimization have investigated diverse facets of optimization methods and their applications. \cite{szks21} investigated the behavior of the mechanism of single-head attention for Seq2Seq model learning, providing insights into how to choose parameters for better performance. \cite{zkv+20} emphasized the importance of adaptive methods for attention models and proposed a new adaptive method for the attention mechanism. \cite{gms23} studied the convergence of over-parameterized neural networks with exponential activation functions, addressing the over-parametrization problem. \cite{lsz23} proposed an algorithm for regularized exponential regression that runs in input sparsity time and demonstrated its effectiveness on various datasets. Finally, \cite{llr23} provided a thorough clarification of how transformers can learn the ``semantic structure'' to detect the patterns of word co-occurrence, exploring the optimization techniques used in transformers and highlighting their strengths and weaknesses.

\paragraph{Learning in-context}

Research on in-context learners based on transformers has been exploring various aspects of their abilities and mechanisms. As an example, \cite{asa+22} showed that these learners can implicitly perform traditional learning algorithms through updating them continuously with new examples and encoding smaller models within their activations. Another work by \cite{gtlv22} focused on training a model that is under the in-context conditions which are used for learning a class of functions, like the linear functions, aiming to determine whether or not a model that has been given information obtained from specific functions within a class can learn the ``majority" of functions in this class through training. In their research, \cite{onr+22} described how Transformers operate as in-context learners and revealed similarities between a few meta-learning formulations, which are based on gradient descent, and the training process of Transformers in in-context tasks. In general, these studies provide valuable insights into the abilities and mechanisms of in-context learners based on transformers, which possess the huge potential to significantly improve the applications of machine learning. \cite{lsx+23} proved a theoretical result about the in-context learning under softmax regression formulation \citep{dls23}.

\paragraph{Fast Attention Computation}

The computation of attention has been explored in several works, with a focus on both dynamic and static attention. \cite{bsz23} investigated the dynamic version of attention computation, where the input data is very dynamic and subject to constant changes, showing both positive results and negative results. They utilized lazy update techniques in their algorithmic results while the hardness result was based on the Hinted MV conjecture. On the other hand, \cite{zhdk23} and \cite{as23} focused on static attention computation. \cite{as23} proposed an algorithm for static attention and provided a hardness result based on the exponential time hypothesis. Meanwhile, \cite{zhdk23} explored the efficiency of static attention algorithms in various applications. \cite{dms23} investigated the sparsification of feature dimension in attention computation, providing both randomized and deterministic algorithms. \cite{syz23} studies the attention kernel regression problem, which utilizes the mathematical induction to generalize the algorithms of solving regression problems $\min_{x \in \R^d} \| A A^\top A x - y \|_2^2$ and $\min_{x \in \R^d} \| A^\top A A^\top A x - y \|_2^2$ to $\min_{x \in \R^d} \| A(A^\top A)^j x - b \|_2$ and $\min_{x \in \R^d} \| (A^\top A)^j x - b \|_2$ respectively, where $j$ is any arbitrary positive integer. \cite{swyz23} provides an algorithm to solve the exact attention regression problem by using the tensor and support vector machine tricks. Moreover, \cite{sxy23} analyzes the polynomial based attention problem, where the $\exp(x)$ function from Eq.~\eqref{eq:attention} is replaced by the $x^{\beta}$ function, where $\beta \geq 2$. Furthermore, \cite{swy23} combines the softmax regression analyzed in \cite{dls23} and the residual neural network developed in \cite{hzrs16} to study a unified regression problem. \cite{lswy23} proposes a two-layer regression problem, where the inner layer is the ReLU function and the outer layer is the softmax regression studied in \cite{dls23}. Finally, \cite{lls+24} studies the masked version of the attention computation showing that any lower triangular matrices can be decomposed into the convolution basis.

%% file: short_preli.tex
\section{Preliminaries}
\label{sec:short_preli}

In this section, we first introduce basic notations. Then, in Section~\ref{sub:short_preli:def}, we define several functions that we use in later sections; in Section~\ref{sub:short_preli:property}, we present a basic mathematical fact.

\paragraph{Notation}
We use $\Z_+$ to represent a set that contains all positive integers, and we use $n$ to be an arbitrary element in $\Z_+$. We define $[n]$ to be the set, i.e., $[n] := \{1, 2, \ldots, n\}$.

 Let $x\in \R^n$ and $y \in \R^n$ be two vectors. For any $i \in [n]$, we let $x_i \in \R$ denote the $i$ -th entry of $x$.  We use $x \circ y \in \R^n$ to represent the vector satisfying $(x\circ y)_i = x_i y_i$ for each $i \in [n]$. We use $\|x\|_p$ (where $p \in \{1,2, \infty\}$) to represent the $\ell_p$ norm of $x$, where $\|x\|_1 := \sum_{i=1}^n |x_i|$ ($\ell_1$ norm), $\|x\|_2 := (\sum_{i=1}^n x_i^2)^{1/2}$ ($\ell_2$ norm), and $\|x\|_\infty := \max_{i\in [n]} |x_i|$ ($\ell_\infty$ norm). For a scalar $z \in \R$, we let $\exp(z)$ represent the standard exponential function.

Note that $\cosh(z) = \frac{1}{2}(\exp(z) + \exp(-z))$ and $\sinh(z) = \frac{1}{2} ( \exp(z)- \exp(-z) )$. Therefore, by the definitions of 
$\exp(z)$, $\cosh(z)$, and $\sinh(z)$, we have $\exp(z)' = \exp(z), \cosh(z)' = \sinh(z), \sinh(z)' = \cosh(z)$
 and 
\begin{align*}
\exp(z)'' &= \exp(z), \\
\cosh(z)'' &= \cosh(z), \\
\sinh(z)'' &= \sinh(z).
 \end{align*}
 
 For an arbitrary vector $x \in \R^n$, we use $\exp(x) \in \R^n$ to denote a vector whose $i$-th entry $\exp(x)_i$ is $\exp(x_i)$. We use $\langle x,y \rangle$ to denote $\sum_{i=1}^n x_i y_i$.  
${ \bf 1}_n$ represents a $n$-dimensional vector whose entries are all $1$, and ${ \bf 0}_n$ represents a $n$-dimensional vector whose entries are all $0$. We use $I_n$ to denote an identity matrix that has size $n \times n$ and all the diagonal are ones.

For an arbitrary vector $u \in \R^n$, let $\diag(u) \in \R^{n \times n}$ represent a diagonal matrix whose $i$-th entry on the diagonal is $u_i$.  
For an arbitrary symmetric matrix $B \in \R^{n \times n}$, we say $B$ is positive definite ($B \succ 0$) if for all vectors $x \in \R^n \backslash \{ {\bf 0}_n \}$, $x^\top B x > 0$. 
For a symmetric matrix $B \in \R^{n \times n}$, we say $B$ is positive semidefinite ($B \succeq 0$) if for all vectors $x \in \R^n$, $x^\top B x \geq 0$.
For symmetric matrices $B$ and $C$, we say $B \succeq C$ if for all $x$, $x^\top B x \geq x^\top C x$.
For any matrix $A$, we use $\| A \|$ to denote the spectral norm of $A$, i.e., $\|A\| = \max_{\|x\|_2 = 1} \|Ax\|_2$. For each $i \in [d]$, we use $A_{*,i} \in \R^n$ to denote the $i$-th column of matrix $A \in \R^{n \times d}$.

\subsection{General Functions: Definitions}
\label{sub:short_preli:def}

In this section, we present the definitions of the basic functions appearing in our loss function.

\begin{definition}\label{def:u}
Let $u(x)$ be one of the following 
\begin{itemize}
    \item Case 1. $u(x) = \exp(Ax)$
    \item Case 2. $u(x) = \cosh(Ax)$
    \item Case 3. $u(x) = \sinh(Ax)$
\end{itemize}
\end{definition}

We define a helpful function as follows.
\begin{definition}\label{def:v}
Let $v(x)$ be one of the following 
\begin{itemize}
    \item Case 1. $v(x) = \exp(Ax)$ (when $u(x) = \exp(Ax)$)
    \item Case 2. $v(x) = \sinh(Ax)$ (when $u(x) = \cosh(Ax)$)
    \item Case 3. $v(x) = \cosh(Ax)$ (when $u(x) = \sinh(Ax)$)
\end{itemize}
\end{definition}
In the above two definitions, we introduce two basic notations $u(x)$ and $v(x)$. Those two notations are utilized in various locations, especially when we compute first derivatives and second derivatives. Note that $x \in \R^d$ is a vector. Therefore, we expect to use $v(x)$ to express a certain part of the derivative of $u(x)$ to simplify our mathematical expression.
 
We define $L_u$ in the following sense: 
\begin{definition}[Loss function $L_u$]\label{def:L_u}
Given a matrix $A \in \R^{n \times d}$ and a vector $b \in \R^n$. 
We define loss function $L_u : \R^d \rightarrow \R$ as  
\begin{align*}
L_{u}(x) := 0.5 \cdot \|  u(x) - \langle u(x) , {\bf 1}_n \rangle \cdot b \|_2^2.
\end{align*}
\end{definition}

For convenience, we define two helpful functions $\alpha$ and $c$:
\begin{definition}[Rescaled coefficients]\label{def:alpha}
Given a matrix $A \in \R^{n \times d}$, we define $\alpha : \R^d \rightarrow \R$ as
    $
    \alpha(x) := \langle u(x), {\bf 1}_n \rangle.
    $
    Then, the $L_u(x)$ (see Definition~\ref{def:L_u}) can be rewritten as 
        $L_u(x) = 0.5 \cdot \| u(x) - b \cdot \alpha(x) \|_2^2$.
\end{definition}

\begin{definition}\label{def:c}
Given a matrix $A \in \R^{n \times d}$ and a vector $b \in \R^n$. Let $\alpha(x)$ be defined as in Definition~\ref{def:alpha}. We define function $c: \R^d \rightarrow \R^n$  as follows
    $
    c(x) := u(x) - b \cdot \alpha(x).
    $ 
    Then, we can rewrite $L_{u}(x)$ (see Definition~\ref{def:L_u}) as
        $L_{u}(x) = 0.5 \cdot \| c(x) \|_2^2$.
\end{definition}

Now, we define the regularization function, $L_{\reg}$.

\begin{definition}\label{def:L_reg:informal}
Given a matrix $A \in \R^{ n \times d}$ and $W = \diag (w) \in \R^{n \times n} $ where $w \in \R^n$ is a vector, we define $L_{\reg} : \R^d \rightarrow \R$ as
\begin{align*}
    L_{\reg} (x) := 0.5 \| W A x\|_2^2  .
\end{align*} 
\end{definition}

\subsection{A Basic Mathematical Property}
\label{sub:short_preli:property}

In this section, we present a basic mathematical property that is useful in later analysis. The following fact provides upper bounds on the norms of exponential, hyperbolic cosine, and hyperbolic sine functions and also establishes an approximation property when input values of these functions are close to each other.

\begin{fact}[Informal version of Fact~\ref{fac:exp_cosh_sinh_vector}]\label{fac:exp_cosh_sinh_vector_informal}
For vectors $a, b \in \R^n$, we have the following results:
\begin{itemize}
    \item $\| \exp(a) \|_{\infty} \leq \exp(\| a \|_2)$
    \item $\| \cosh(a) \|_{\infty} \leq \cosh( \| a \|_2 ) \leq \exp( \| a \|_2 )$
    \item $\| \sinh(a) \|_{\infty} \leq \sinh(\| a \|_2) \leq \cosh(\| a \|_2) \leq \exp(\| a \|_2)$
    \item $\cosh(a) \circ \cosh(a) - \sinh(a) \circ \sinh(a) ={\bf 1}_n$
\end{itemize}
Approximation in a small range: If two vectors $a, b \in \R^n$ are close, meaning $\| a - b \|_{\infty} \leq 0.01$, then, we can get
\begin{itemize}
    \item $\| \exp(a) - \exp(b) \|_2 \leq \| \exp(a) \|_2 \cdot 2 \| a - b \|_{\infty}$,
    \item $\| \cosh(a) - \cosh(b) \|_2 \leq \| \cosh(a) \|_2 \cdot 2 \| a - b \|_{\infty}$, and
    \item $\| \sinh(a) - \sinh(b) \|_2 \leq \| \cosh(a) \|_2 \cdot 2 \| a - b \|_{\infty}$.
\end{itemize}
\end{fact}

This fact shows that the three distinct functions—exponential, hyperbolic cosine, and hyperbolic sine—actually share some similar mathematical properties.

%% file: tech.tex
\section{Technique Overview}
\label{sec:technique_overview}

An overview of our techniques is presented in this section.

\paragraph{General Functions}

For the purpose of applying our theory to $\exp$, $\sinh$, and $\cosh$ functions at the same time, we will introduce our generalized definition first. $u(x)$ is used to represent the functions including $\exp, \sinh$ and $\cosh$. With the aim that we can only use $u(x)$ in the following proof, the common property used in our proof of $u(x)$ will be proposed. To elaborate further, the expression for $u(x)$ is defined as Definition~\ref{def:u}.
Based on Fact~\ref{fac:exp_cosh_sinh_vector_informal} and $\| A \| \leq R$, we have 
\begin{align*}
    \| u(x) \|_2 \leq \sqrt{n} \exp{(R^2)}
\end{align*}
    
And $v(x)$ is as defined as Definition~\ref{def:v}. A unified version of the Hessian computation and the gradient computation can also be obtained as follows:
\begin{itemize}
    \item $\frac{\d u(x)}{\d x} = ( v(x) {\bf 1}_d^\top ) \circ A$
    \begin{itemize}
        \item $\frac{\d u(x)}{\d x_i}= v(x) \circ A_{*,i}$ for each $i \in [d]$
    \end{itemize}
    \item $\frac{\d^2 u(x)}{ \d x_i^2} = A_{*,i} \circ u(x) \circ A_{*,i}$ for each $i \in [d]$
    \item $\frac{\d^2 u(x)}{\d x_i \d x_j} = A_{*,i} \circ u(x) \circ A_{*,j}$ for each $i,j \in [d] \times [d]$
\end{itemize}
{\bf Hessian Computation}
Taking $w \in \R^d$ into account as well, the target function we are focusing on is listed as follows: 
\begin{align}\label{eq:target_function}
    \min_{x \in \R^d} L(x) 
    = & ~ \min_{x \in \R^d} \underbrace{0.5 \cdot \| u(x) - \langle  u(x) , {\bf 1}_n \rangle \cdot b \|_2^2}_{:= L_{u}} \notag\\
    + & ~ \underbrace{0.5 \cdot \| \diag(w) A x\|_2^2}_{:= L_{\reg}}.
\end{align}
The computation of the Hessian for the problem directly is complex. We will introduce some techniques used in the computation of the Hessian function. 
To enhance the clarity of our expression, we draw a comparison between our Hessian Computation and the ones presented in \cite{lsz23, dls23}. Specifically, we introduce the function $\alpha(x) := \langle u(x), {\bf 1}_n \rangle$ and note that in \cite{lsz23}, there is no need to compute $\alpha(x)$, while $\alpha^{-1}(x)$ is the focus of \cite{dls23}. However, our emphasis is on the function $\alpha(x)$.

Additionally, with the definition $c(x) := u(x) - b \cdot \alpha(x)$, the computation of the Hessian can be divided into the $\frac{\d^2 u(x)}{\d x^2}$, $\frac{\d^2 \alpha(x)}{\d x^2}$ and $\frac{\d^2 c(x)}{\d x^2}$.

\paragraph{$\frac{\d^2 L}{\d x^2}$ is Positive Definite}
The first property we need to establish in order to apply the Newton optimization method is the positive definiteness of the Hessian. This is inspired by the semidefinite programming literature \citep{a00,hjs+22}. We have defined $R_0 := \max\{ \| v(x) \|_2, \| b \|_2, \| c(x) \|_2, \| u (x) \|_2, 1\}$. Give that 
\begin{align*}
\frac{\d^2 L(x) }{ \d x_i^2} = \underbrace{A_{*,i}^\top B(x) A_{*,i}}_{\frac{\d^2 L_{u}(x) }{ \d x_i^2}} + \underbrace{A_{*,i}^\top W^2 A_{*,i}}_{\frac{\d^2 L_{\reg}(x) }{ \d x_i^2}}
\end{align*}
and the bound on $B(x)$
\begin{align*}
    -10 R_0^4 \cdot I_n \preceq B(x) \preceq 10 R_0^4 \cdot I_n,
\end{align*} 
by choosing $w_i \geq 10 R_0^4 + l/\sigma_{\min} (A)^2$, the Hessian function is Positive definite now (see Section~\ref{sec:PSD} for detail).

\paragraph{$\frac{\d^2 L}{\d x^2}$ is Lipschitz with respect to variable $x$}
To apply the Newton optimization method, it is also necessary to ensure the Lipschitz property. To finish the proof, we will get the upper bound of $\| H(x) - H(y) \|$ by $c \cdot \| x - y\|_2$ where $c$ is a scalar. $H(x)$ can be decomposed into $G_i$ where $i \in [n]$. 
\begin{align*}
     \| H(x) - H(y) \|
    \leq & ~ \| A \| \cdot (\sum_{i=1}^5 \| G_i \|) \| A \| 
\end{align*}
The idea of how to bound each term $G_i$ is quite standard neural network literature (for example, see \cite{als19_rnn,als19_dnn}).

With 
\begin{align*}
    R_{\infty}
    := & ~ \max \{ \| u(x) \|_2, \| u (y) \|_2, \| v(x) \|_2,\\
    & ~ \| v(y)\|_2, \| c(x) \|_2, \| c(y) \|_2, \| b \|_2, 1 \}
\end{align*}
and then we get the following bound on $\| H(x) - H(y) \|$ by the following equation:
\begin{align*}
    & ~ \sum_{i=1}^5 \| G_i \| \\
    \leq & ~  20 R_{\infty}^3 \cdot \max\{ \| u(x) - u(y) \|_2 , \| c(x) - c(y) \|_2 \}.
\end{align*}
Furthermore, we can prove that the Hessian is Lipschitz continuous $\| H(x) - H(y) \| \leq  ~ n^{4} \exp(20 R^2) \cdot \| x - y \|_2$ (see details in Section~\ref{sec:lipschitz}).

\paragraph{Approximated Newton Algorithm}

Based on the property of the Hessian function we have, we can apply the approximated Newton Method to the function regression problem. 
Building on the assumption of a $(l, M)$-good loss function, we can guarantee the correctness of our algorithm.

Given $M = n^{4} \exp(20 R^2)$, $x_*$ as the optimization of Eq.~\eqref{eq:target_function} and $x_0$ as the initialization, we have a good initialization assumption 
\begin{align*}
    \underbrace{ \| x_0 - x_* \|}_{:= r_0} M \leq 0.1 l
\end{align*}
To expedite the algorithm computation, it is natural to introduce a method for approximating the Hessian and its inverse (for example, see \cite{cls19,lsz19,song19_thesis,b20,jswz21,sy21,hjs+22,gs22,dsw22,syyz22,jlsz23}). Given that $H(x_t)$ is a diagonal matrix, $\frac{\d^2 L}{\d x^2}$ can be transformed into a format $A^\top H(x_t) A$. With $\epsilon_0 \in (0,0.1)$, an alternative way to obtain a sparse method is to substitute $H(x_t)$ with a sparse matrix $\wt{H}(x_t)$ where
\begin{align*}
     (1-\epsilon_0) \cdot H(x_t) \preceq \wt{H}(x_t) \preceq (1+\epsilon_0) \cdot H(x_t)
\end{align*}
The running time of Hessian computation can be reduced to $\wt{O}(\nnz(A)+d^{\omega})$. To ensure the convergence of our algorithm, the number of iterations is expected to be $\log(1/\epsilon)$ based on the assumption above, leading to a total running time of 
\begin{align*}
\wt{O}((\nnz(A)+d^{\omega})\cdot \log(1/\epsilon).    
\end{align*}
Here $\nnz(A)$ denotes the number of nonzero entries in matrix $A$. 
Thus, we can derive our main result Theorem~\ref{thm:main_informal}.

\paragraph{From Lipschitz with respect to $x$ to Lipschitz with to $A$}

In Section~\ref{sec:lipschitz}, we already proved a number of results for Lipschitz with respect to $x$. To present the application to in-context learning for rescaled softmax regression, we generalize the Lipschitz with respect $x$ to Lipschitz with respect to $A$ (see Setion~\ref{sec:lipschitz:A}). 
To analyze the Lipschitz property, we bound $\| c(A) - c(B) \|_2$ using two terms $\| u(A) - u(B) \|_2 $ and $ | \alpha(A) - \alpha(B) |$.

Let $u(x)$ be in Definiton~\ref{def:u} and $u(A) = \exp(Ax)$, we have
\begin{align*}
    \| u(A) - u(B) \|_2 \leq 2 \sqrt{n} R \exp(R^2) \| A - B \|
\end{align*}
We can also have
\begin{align*}
    | \alpha(A) - \alpha(B) | \leq \sqrt{n} \cdot \| u(A) - u(B) \|_2
\end{align*}
Then $\| c(A) - c(B) \|_2$ can be bounded as follows
\begin{align*}
    \| c(A) - c(B) \|_2 
    \leq & ~ \| u (A) - u(B) \|_2 \\
    + & ~ | \alpha(A) -  \alpha(B) | \cdot \| b \|_2.
\end{align*}
The Lipschitz property of $c(A)$ with respect to $A$ is guaranteed by $\| c(A) - c(B) \|_2 \leq C \|A - B \|$, where $C$ is a scalar that can be determined as described above. Finally, we present the Corollary~\ref{cor:in_context_learning} as our in-context learning application.

%% file: short_hessian.tex
\section{Properties of Hessian}\label{sec:short_hessian}

In this section, we introduce and analyze two crucial components of our main result (see Theorem~\ref{thm:main_formal}). In Section~\ref{sub:short_hessian:bound}, we show that Hessian is a positive definite matrix. In Section~\ref{sub:short_hessian:Lipschitz}, we analyze the Lipschitz property of Hessian. Both of the properties are the promise of correctness and efficiency of our Algorithm~\ref{alg:main:informal}.

\subsection{Hessian is Positive Definite}
\label{sub:short_hessian:bound}
In this section, we present the result that Hessian is positive definite, which is the promise in computing $\wt{H}$ efficiently (see Lemma~\ref{lem:subsample}). Due to space limitation, we only present the informal Lemma statement here and defer the formal Lemma statement and the proof to Section~\ref{sub:PSD:lower_bound}.

\begin{lemma}[Informal version of Lemma~\ref{lem:hessian_psd_exp}]\label{lem:hessian_psd_exp:informal}

Let $ A \in \mathbb{R}^{n \times d} $, where $ u(x) $ is defined according to Definition~\ref{def:u}, and $ v(x) $ follows Definition~\ref{def:v}. Furthermore, $ L_{u}(x) $ is defined as per Definition~\ref{def:L_u}, and $ L_{\reg}(x) $ corresponds to Definition~\ref{def:L_reg:informal}. The combined loss function is denoted as 
\begin{align*}
    L(x) = L_{\reg}(x) + L_{u}(x).
\end{align*}

Given a vector $ w \in \mathbb{R}^n $, the diagonal matrix $ W = \diag(w) \in \mathbb{R}^{n \times n} $, and $ W^2 $ represents the matrix with $ w_i^2 $ as the $ i $-th diagonal entry. Here, $ \sigma_{\min}(A) $ denotes the minimum singular value of $ A $, and $ l > 0 $ is a scalar. Let $R_0 = \max\{ \| u (x) \|_2, \| v \|_2, \| b \|_2, \| c(x) \|_2, 1\}$. Suppose for all $i\in [n]$, $w_i^2 \geq 10 R_0^4 + l/\sigma_{\min} (A)^2$.

Then we have 
\begin{align*}
    \frac{\d^2 L(x)}{ \d x^2} \succeq l \cdot I_d.
\end{align*}
\end{lemma}

\subsection{Hessian is Lipschitz}
\label{sub:short_hessian:Lipschitz}

In the following lemma, we show that the Hessian is Lipschitz, which is used to demonstrate that our loss function is $(l,M)$-good (see Definition~\ref{def:assumptions}). The proof is deferred to Section~\ref{sec:lipschitz}.
\begin{lemma}[Informal version of Lemma~\ref{lem:hessian_lipschitz}]\label{lem:hessian_lipschitz:informal}
Let $H(x) = \frac{ \d^2 L}{\d x^2} $ and $R > 4$. Let $\| x \|_2 \leq R$, $\| y \|_2 \leq R$, where $x, y \in \R^d$. Let $\| A (x-y) \|_\infty < 0.01$, where $A \in \R^{n \times d}$, $\| A \| \leq R$, $\| b \|_2 \leq R$, where $b \in \R^n$, and
\begin{align*}
        R_{\infty}
        := & ~ \max\{ \|u(x) \|_2, \| u(y) \|_2, \| c(x) \|_2, \| c(y) \|_2, 1 \} \\
        \leq & ~ 2n R \exp(R^2).
    \end{align*}

Then we have 
\begin{align*}
\| H(x) - H(y)\| \leq  n^{4} \exp (20 R^2) \cdot \| x - y \|_2
\end{align*}
\end{lemma}

%% file: short_newton.tex
\section{Regression Algorithm}\label{sec:newton:short}

\begin{algorithm}[!ht]\caption{Rescaled Hyperbolic Functions Regression.}\label{alg:main:informal}
\begin{algorithmic}[1]
\Procedure{RescaledHyperbolicFunctionsRegression}{$A \in \R^{n \times d},b \in \R^n,w \in \R^n, \epsilon, \delta$} \Comment{Theorem~\ref{thm:main_formal}} 
    \State We choose $x_0$ (suppose it satisfies Definition~\ref{def:assumptions})  
    \State We use $T \gets \log( \| x_0 - x^* \|_2 / \epsilon )$ to denote the number of iterations.  
    \For{$t=0 \to T$} 
        \State $D \gets B_{\diag}(x_t) + \diag(w \circ w)$ 
        \State $\wt{D} \gets \textsc{SubSample}(D,A,\epsilon_1 = \Theta(1), \delta_1 = \delta/T)$ \Comment{Lemma~\ref{lem:subsample}}
        \State $g \gets  A^\top ( c(x_t) \circ v(x_t) - v(x_t) \langle b, c(x_t) \rangle )  + A^\top W^2 A x_t$
        \Comment{Definition~\ref{def:v} and Definition~\ref{def:c}}
        \State $\wt{H} \gets A^\top \wt{D} A$ \label{line:approximate_H}
        \State $x_{t+1} \gets x_t - \wt{H}^{-1} g$ \label{line:update}
        \Comment{Definition~\ref{def:update_x_t}}
    \EndFor
    \State $\wt{x}\gets x_{T+1}$
    \State \Return $\wt{x}$
\EndProcedure
\end{algorithmic}
\end{algorithm}

We provide an overview of our algorithm and its key components in this section. To help readers better understand our contribution and how it relates to the results in Section~\ref{sec:short_hessian}, we explain some of the key parts of our algorithm in this section. Given that $L(x) = L_{u}(x) + L_{\reg}(x)$, 
we consider the following optimization problem $\min_{x \in \R^d} L(x)$.

Specifically, in Section~\ref{sub:newton:good:short}, we introduce the $(l,M)$-good Loss function. In Section~\ref{sub:newton:approximation:short}, we present the approximate Hessian and update rule.

\subsection{\texorpdfstring{$(l,M)$}{}-good Loss function}
\label{sub:newton:good:short}

In this section, we explain the meaning of $(l,M)$-good loss function used in Lemma~\ref{lem:one_step_shrinking}. Now, we provide the following definition: 
\begin{definition}[$(l,M)$-good Loss function]\label{def:assumptions}
For a function $L : \R^d \rightarrow \R$, we say $L$ is $(l,M)$-good if satisfies the following conditions,
\begin{itemize}
    \item {\bf $l$-local Minimum.}  
    We define $l >0$ to be a positive scalar. If there exists a vector $x^* \in \R^d$ such that 
    $\nabla L(x^*) = {\bf 0}_d$ and $\nabla^2 L(x^*) \succeq l \cdot I_d$.
    \item {\bf Hessian is $M$-Lipschitz.} If there exists a positive scalar $M>0$ such that
    $ \| \nabla^2 L(y) - \nabla^2 L(x) \| \leq M \cdot \| y - x \|_2 $
    \item {\bf Good Initialization Point.} Let $x_0$ denote the initialization point. If $r_0:=\| x_0 -x_*\|_2$ satisfies
    $ r_0 M \leq 0.1 l.$
\end{itemize}
\end{definition}
Based on Lemma~\ref{lem:hessian_lipschitz:informal}, our loss function (see Definition~\ref{def:rescaled_softmax_regression}) satisfies the $(l,M)$-good assumption above.
Now, we turn to two key steps in our main Algorithm~\ref{alg:main:informal}: Line~\ref{line:approximate_H} and Line~\ref{line:update}.

\subsection{Approximate of Hessian and Update Rule}\label{sub:newton:approximation:short}

In this section, we present the concept of approximate update and its properties. The approximate update replaces the Hessian matrix $H(x_t) \in \R^{d \times d}$ in the well-known Newton method $x_{t+1} = x_t - H(x_t)^{-1} \cdot g(x_t)$ by the approximate Hessian $\wt{H}(x_t) \in \R^{d \times d}$, which is close to $H(x_t)$. The formal definition of the approximate Hessian is presented as follows:
\begin{definition}[$\epsilon_0$-approximate Hessian]\label{def:wt_H}
Let $x \in \R^d$ and $H(x) \in \R^{d \times d}$ be a Hessian matrix. For all $\epsilon_0 \in (0, 0.1)$, we define an $\epsilon_0$-approximate Hessian\footnote{This approximate Hessian does not need to be a Hessian matrix. It is used to approximate the Hessian $H(x) \in \R^{d \times d}$.} $\wt{H}(x) \in \R^{d \times d}$ to be a matrix that satisfies:
\begin{align*}
 (1-\epsilon_0) \cdot H(x) \preceq \wt{H}(x) \preceq (1+\epsilon_0) \cdot H(x).
\end{align*}
\end{definition}

Using the definition of the approximate Hessian, we define the approximate Newton method as follows:
\begin{definition}[$\epsilon_0$-approximate update Newton's method \citep{dls23}]\label{def:update_x_t}

Let $L: \R^d \rightarrow \R$ be a loss function. Suppose it has the gradient function $g: \R^d \rightarrow \R^d$ and the Hessian matrix $H : \R^d \rightarrow \R^{d \times d}$. Let $\wt{H} : \R^d \rightarrow \R^{d \times d}$ be an $\epsilon_0$-approximate Hessian defined in Definition~\ref{def:wt_H}, for any $\epsilon_0 \in (0, 0.1)$. An $\epsilon_0$-approximate update of Newton's method is a recurrence relation defined on $L$:
\begin{align*}
    x_{t+1} = x_t  - \wt{H}(x_t)^{-1} \cdot  g(x_t)  .
\end{align*}
\end{definition}

In Line~\ref{line:approximate_H}, we need to compute an approximated $\wt{H}$ (see Definition~\ref{def:wt_H}).
In order to get the approximated Hessian $\wt{H}(x_t)$ efficiently, we present a standard tool that can be found in Lemma~4.5 of \cite{dsw22}.
\begin{lemma}[\citep{dsw22,syyz22}]\label{lem:subsample}
Let $\epsilon_0, \delta \in (0, 0.1)$ be the precision parameter and failure probability, respectively. 
Let $A \in \R^{n \times d}$. 

Then, for all $i \in [n]$, for all $D \in \R^{n \times n}$ satisfying $D_{i, i} > 0$, there exists an algorithm which runs in time
\begin{align*}
O( (\nnz(A) + d^{\omega} ) \poly(\log(n/\delta)) )
\end{align*}
and outputs an $O(d \log(n/\delta))$ sparse diagonal matrix $\wt{D} \in \R^{n \times n}$, i.e. a diagonal matrix where most of the entries are zeros, and the number of non-zero entries is less than or equal to a constant time $d \log(n/\delta)$, such that 
\begin{align*}
(1- \epsilon_0) A^\top D A \preceq A^\top \wt{D} A \preceq (1+\epsilon_0) A^\top D A.
\end{align*}
Here $\omega$ denotes exponent of matrix multiplication, currently $\omega \approx 2.373$ \citep{w12,aw21}.
\end{lemma}

Given the importance of the approximated Hessian computation in the update step (see Line~\ref{line:update}), we now focus on this particular step of Algorithm~\ref{alg:main:informal}, where $x_{t+1} = x_t - \wt{H}(x_t)^{-1} \cdot g(x_t)$. To establish the correctness of our algorithm, we leverage Lemma~6.9 of \cite{lsz23}:
\begin{lemma}[Iterative shrinking, Lemma 6.9 on page 32 of \cite{lsz23}]\label{lem:one_step_shrinking}
For a positive integer $t$, we define $x_t \in \R^d$ to be the $t$-th iteration of the approximate Newton method (see Definition~\ref{def:update_x_t}). We let $x^* \in \R^d$ be defined as in Definition~\ref{def:assumptions}, for fixed $A \in \R^{n \times d}$, $b \in \R^n$, and $w \in \R^n$. 
Let $L: \R^d \to \R$ be a loss function which is $(l,M)$-good (see Definition~\ref{def:assumptions}). 
Let $r_t:= \| x_t - x^* \|_2$. Let $\ov{r}_t: = M \cdot r_t$.

Then, for all $\epsilon_0 \in (0,0.1)$, we have  
\begin{align*}
r_{t+1} \leq 2 \cdot (\epsilon_0 + \ov{r}_t/( l - \ov{r}_t ) ) \cdot r_t.
\end{align*} 
\end{lemma}

This lemma allows us to shrink the distance $\| x_t - x^* \|_2$  by one step using our assumption that the loss function is $(l,M)$-good, as verified in Section~\ref{sec:short_hessian}. 
To apply Lemma~\ref{lem:one_step_shrinking}, we need the following induction hypothesis lemma. This is very standard in the literature, see \cite{lsz23}.

\begin{lemma}[Induction hypothesis, Lemma 6.10 on page 34 of \cite{lsz23}]\label{lem:newton_induction}
For a positive integer $t$, for each $i \in [t]$, we define $x_i \in \R^d$ to be the $i$-th iteration of the approximate Newton method (see Definition~\ref{def:update_x_t}). We let $x^* \in \R^d$ be defined as in Definition~\ref{def:assumptions}, for fixed $A \in \R^{n \times d}$, $b \in \R^n$, and $w \in \R^n$.  For each $i \in [t]$, we define $r_i:= \| x_i - x^* \|_2$. Let $\epsilon_0 \in (0, 0.1)$. Suppose $r_{i} \leq 0.4 \cdot r_{i-1}$, for all $i \in [t]$. For $M$ and $l$ to be defined for Definition~\ref{def:assumptions}, we assume $M \cdot r_i \leq 0.1 l$, for all $i \in [t]$.

Then we have
\begin{itemize}
    \item $r_{t+1} \leq 0.4 r_t$.
    \item $M \cdot r_{t+1} \leq 0.1 l$.
\end{itemize}
\end{lemma}

By applying this induction hypothesis and choosing a sufficiently large value of the number of iterations, we can then establish the correctness of our algorithm. The running time of our algorithm in each iteration is dominated by $O( (\nnz(A) + d^{\omega} ) \poly(\log(n/\delta)) )$. Because of the page limit, we delay our formal proof to Appendix~\ref{sec:main_result}.

%% file: conclusion.tex
\section{Conclusion}\label{sec:conclusion}

The exponential function-based attention unit is a crucial component in LLMs, enabling the model to selectively focus on different parts of the input sequence and improving its ability to capture long-range dependencies.
In this paper, we focus on a slightly different version of softmax regression, namely
\begin{align*}
    \min_{x \in \mathbb{R}^d } \| u(x) - \langle u(x) , {\bf 1}_n \rangle \cdot b \|_2.
\end{align*}
We propose an efficient algorithm for this problem that operates on sparse inputs, leveraging the positive-definite and Lipschitz properties of the Hessian. 
Our mathematical analysis provides a deeper theoretical understanding of optimization problems related to the attention mechanism in LLMs. 
This could spur further advances and innovations in the architecture and training of language models.

Moreover, our algorithm framework is highly general and can be applied to a variety of functions, including $\exp(\cdot)$, $\cosh(\cdot)$, and $\sinh(\cdot)$.

\section*{Acknowledgments}

This work was mostly done when Junze Yin was at Boston University. Junze Yin is supported by the Rice University graduate fellowship.

%% file: preli.tex
\paragraph{Roadmap.}
We define the notations and propose approximate algebra, differential computation, and math tools for exact algebra used in our paper in Section~\ref{sec:preli}. 
In Section~\ref{sec:g_H}, we introduce the computation of Hessian and Gradient. In Section~\ref{sec:PSD}, we prove $L= L_{u} + L_{\reg}$ is convex function. The hessian of $L= L_{u} + L_{\reg}$ is proved to be Lipschitz in Section~\ref{sec:lipschitz}. In Section~\ref{sec:lipschitz:A}, we analyze the Lipschitz with respect to $A$, where $A \in \R^{n \times d}$. In Section~\ref{sec:main_result}, we introduce our main result and algorithm.

\section{Preliminaries}\label{sec:preli}
In Section~\ref{sub:preli:notations}, we introduce several basic notations and mathematics symbols, which are used in this paper. In Section~\ref{sub:preli:basic_algebra}, we present the algebraic properties for $\circ$ and $\diag$. In Section~\ref{sub:preli:inner_product}, the properties of the inner product are explained. In Section~\ref{sub:preli:positive_semidefinite}, the properties of the $\preceq$ and its relationship with $\diag$ and $\circ$ are introduced. In Section~\ref{sub:preli:calculus}, we present several standard derivative rules, both for the scalar variables and for the vector variables. In Section~\ref{sub:preli:vector}, we demonstrate the properties of the vector norm bound, including the Cauchy-Schwarz inequality and other inequalities of the bound containing $\circ$ and $\diag$. In Section~\ref{sub:preli:matrix}, we illustrate the properties of the matrix norm bound, namely the inequalities of the spectral norms. In Section~\ref{sub:preli:scalar_version}, we introduce the properties of the hyperbolic functions, which take the scalar as an element of their domains. On the other hand, in Section~\ref{sub:preli:vector_version}, we elucidate the properties of the hyperbolic functions, which take the vector as an element of their domains. In Section~\ref{sub:preli:regularization}, we define the  regularization function, $L_{\reg} : \R^d \to \R$, and analyze its basic properties. In Section~\ref{sub:preli:gradient}, we define the gradient and Hessian of an arbitrary loss function $L$ and define the update of the Newton method.

\subsection{Notation}\label{sub:preli:notations}
In this section, we explain the several basic notations.
We use $\Z_+$ to represent a set that contains all the positive integers, and we use $n$ to be an arbitrary element in $\Z_+$. We define $[n]$ to be the set, i.e., $[n] := \{1, 2, \ldots, n\}$.
 Let $x\in \R^n$ and $y \in \R^n$ be two vectors. For any $i \in [n]$, we let $x_i \in \R$ to denote the $i$-th entry of $x$.  We use $x \circ y \in \R^n$ to represent the vector satisfying $(x\circ y)_i = x_i y_i$ for each $i \in [n]$. We use $\|x\|_p$ (where $p \in \{1,2, \infty\}$) to represent the $\ell_p$ norm of $x$, where $\|x\|_1 := \sum_{i=1}^n |x_i|$ ($\ell_1$ norm), $\|x\|_2 := (\sum_{i=1}^n x_i^2)^{1/2}$ ($\ell_2$ norm), and $\|x\|_\infty := \max_{i\in [n]} |x_i|$ ($\ell_\infty$ norm).
 For a scalar $z \in \R$, we let $\exp(z)$ represent the standard exponential function.
 We then define $\cosh(z) := \frac{1}{2}(\exp(z) + \exp(-z))$ and $\sinh(z):=\frac{1}{2} ( \exp(z)- \exp(-z) )$.
 Note that
 \begin{align*}
\exp(z)' = \exp(z), \cosh(z)' = \sinh(z), \sinh(z)' = \cosh(z)
 \end{align*}
 and 
\begin{align*}
\exp(z)'' = \exp(z), \cosh(z)'' = \cosh(z), \sinh(z)'' = \sinh(z).
 \end{align*}
 For an arbitrary vector $x \in \R^n$, we use $\exp(x) \in \R^n$ to denote a vector whose $i$-th entry $\exp(x)_i$ is $\exp(x_i)$. We use $\langle x,y \rangle$ to denote $\sum_{i=1}^n x_i y_i$.  
${ \bf 1}_n$ represents a $n$-dimensional vector whose entries are all $1$, and ${ \bf 0}_n$ represents a $n$-dimensional vector whose entries are all $0$.
For an arbitrary vector $u \in \R^n$, let $\diag(u) \in \R^{n \times n}$ represent a diagonal matrix whose $i$-th entry on diagonal is $u_i$.  
For an arbitrary symmetric matrix $B \in \R^{n \times n}$, we say $B$ is positive definite ($B \succ 0$) if for all vectors $x \in \R^n \backslash \{ {\bf 0}_n \}$, $x^\top B x > 0$. 
For a symmetric matrix $B \in \R^{n \times n}$, we say $B$ is positive semidefinite ($B \succeq 0$) if for all vectors $x \in \R^n$, $x^\top B x \geq 0$.
For symmetric matrices $B$ and $C$, we say $B \succeq C$ if for all $x$, $x^\top B x \geq x^\top C x$.
For any matrix $A$, we use $\| A \|$ to denote the spectral norm of $A$, i.e., $\|A\| = \max_{\|x\|_2 = 1} \|Ax\|_2$. 
For each $i \in [d]$, we use $A_{*,i} \in \R^n$ to denote the $i$-th column of matrix $A \in \R^{n \times d}$.
We use $I_n$ to denote an identity matrix that has size $n \times n$ and all the diagonal are ones.

\subsection{Basic Algebra for \texorpdfstring{$\circ$}{} and \texorpdfstring{$\diag$}{}}
\label{sub:preli:basic_algebra}

In this section, we provide a fact that includes the basic algebraic properties of $\circ$ and $\diag$.

\begin{fact}\label{fac:circ_diag}
Given vectors $a\in \R^n$, $b \in \R^n $, $c \in \R^n$ and $d \in \R^n$,  we have
\begin{itemize}
    \item $a \circ b = \diag(a) \cdot b =  \diag(a) \cdot \diag(b) \cdot {\bf 1}_n$
    \begin{itemize}
        \item $a \circ b = b \circ a$
        \item $\diag(a) b = \diag(b) a$
        \item $\diag (a) \cdot \diag (b) \cdot {\bf 1}_n = \diag(b) \cdot \diag(a) \cdot {\bf 1}_n$
    \end{itemize}
    \item $\diag ( a \circ b ) = \diag (a) \diag (b)$
    \item $\diag(a)+\diag(b)=\diag(a+b)$
    \item $a^{\top} (b \circ c) = a^{\top} \diag(b) c$
    \begin{itemize}
        \item $a^{\top} (b \circ c) = b^{\top} (a \circ c)= c^{\top} (a \circ b)$
        \item $a^\top \diag(b) c = b^\top \diag(a) c = a^\top \diag(c) b$
    \end{itemize}
    \item $\langle a, b\circ c\rangle = a^\top \diag(b) c$
\end{itemize}
\end{fact}

\subsection{Basic Inner Product}
\label{sub:preli:inner_product}

Now, we present the inner product properties.

\begin{fact}\label{fac:inner_product}
Given vectors $a\in \R^n$, $b \in \R^n $ and $c \in \R^n$,  we have
\begin{itemize}
    \item $\langle a,b \rangle = \langle b,a \rangle$
    \item $\langle a \circ b, c \rangle = \langle a \circ b \circ c,  {\bf 1}_n \rangle$
    \item $\langle a,b \rangle = a^\top b= b^\top a$
    \item $\langle a, b \rangle = \langle a \circ b, {\bf 1}_n \rangle$
    \item $\| a - b \|_2^2 = \| a \|_2^2 + \| b \|_2^2 - 2 \langle a , b \rangle$
    \item $\langle a, b \rangle \langle c, d \rangle = a^\top b c d^\top = b^\top a c d^\top $
\end{itemize}
\end{fact}

\subsection{Positive Semi-definite}
\label{sub:preli:positive_semidefinite}

In this section, we explain the properties of the mathematics operation $\preceq$.

\begin{fact}\label{fac:psd}
    Let $u, v \in \R^n$. 
    We have:
    \begin{itemize}
        \item $u u^\top \preceq \|u\|_2^2 \cdot I_n$. 
        \item $\diag(u) \preceq \|u\|_2 \cdot I_n$
        \item $\diag(u \circ u) \preceq \|u\|_2^2 \cdot I_n$
       \item $\diag(u \circ v) \preceq \|u\|_2 \cdot \|v\|_2 \cdot I_n$
        \item $uv^\top + v u^\top \preceq u u^\top + v v^\top$
        \item $uv^\top + v u^\top \succeq -( u u^\top + v v^\top )$
        \item $(v \circ u) (v \circ u )^\top \preceq \| v \|_{\infty}^2 u u^\top$
        \item $(v \circ u) u^\top \preceq \| v\|_{\infty} u u^\top$
        \item  $(v \circ u) u^\top \succeq - \| v\|_{\infty} u u^\top$
    \end{itemize}
\end{fact}

\subsection{Basic Calculus and Chain Rule}
\label{sub:preli:calculus}

In this section, we present the basic calculus rules, including the derivative rules for scalars and the derivative rules for vectors.

\begin{fact}\label{fac:chain_rule}
We have
\begin{itemize}
    \item Let $\alpha \in \R$ be a fixed scalar, let $x \in \R^d$ denote variable, then we have $\frac{\d \alpha x}{\d t} = \alpha \frac{\d x}{\d t} $
    \item Let $f(x) \in \R^n$, we have $\frac{\d 0.5 \| f(x) \|_2^2}{ \d t} = \langle f(x), \frac{\d f(x)}{ \d t} \rangle $
    \item Let $b \in \R^n$ be a fixed vector, we have $\frac{\d (b \circ f(x))}{\d t} = b \circ \frac{\d f(x)}{ \d t}$  
    \item Let $z \in \R$ denote a scalar variable, we have the following calculus rules
    \begin{itemize}
        \item $\frac{\d \exp(z)}{ \d t} = \exp(z) \frac{\d z}{\d t}$
        \item $\frac{\d \cosh(z)}{\d t} = \sinh(z) \frac{\d z}{\d t}$
        \item $\frac{\d \sinh(z)}{\d t} = \cosh(z) \frac{\d z}{\d t}$
    \end{itemize}
    \item Let $x \in \R^n$ denote a vector variable, we have the following calculus rules
    \begin{itemize}
        \item $\frac{\d \exp(x)}{\d t} = \exp(x) \circ \frac{\d x}{ \d t}$
        \item $\frac{\d \cosh(x)}{\d t} = \sinh(x) \circ \frac{\d x}{ \d t}$
        \item $\frac{\d \sinh(x)}{\d t} = \cosh(x) \circ \frac{\d x}{ \d t}$
    \end{itemize}
\end{itemize}
\end{fact}

\subsection{Basic Vector Norm Bounds}
\label{sub:preli:vector}

Now, we analyze the norm bounds for vectors.

\begin{fact}\label{fac:basic_vector_norm}
Given vectors $a \in \R^n$ and $b \in \R^n$, we have
\begin{itemize}
    \item $\langle a,b \rangle \leq \| a \|_2 \cdot \|b\|_2$ (Cauchy-Schwarz inequality)
    \item $\| \diag (a) \| \leq \| a \|_\infty \leq \| a \|_2$
    \item $\| a \circ b \|_2 \leq \| a \|_\infty \cdot \| b \|_2 \leq \| a \|_2 \cdot \| b\|_2 $
    \item $\| a \|_{\infty} \leq \| a \|_2 \leq \sqrt{n} \cdot  \| a \|_\infty$ ($\ell_{\infty}$-norm vs $\ell_2$-norm)
    \item $\| a \|_2 \leq \| a \|_1 \leq \sqrt{n} \cdot \| a \|_2$ ($\ell_2$-norm vs $\ell_1$-norm)
    
\end{itemize}
\end{fact}

\subsection{Basic Matrix Norm Bound}
\label{sub:preli:matrix}

Then, we analyze the norm bounds for matrices. 

\begin{fact}\label{fac:basic_matrix_norm}
For matrices $A$ and $B$, we have
\begin{itemize}
    \item Let $a,b \in \R^d$ denote two vectors, then we have $\| a b^\top \| \leq  \| a \|_2 \cdot \| b \|_2$. 
    \item $\| A x \| \leq \| A \| \cdot \| x \|_2$.
    \item $\| A B \| \leq \| A \| \cdot \| B \|$
    \item Let $\alpha \in \R$ be a scalar, then we have $\| \alpha \cdot A \| \leq |\alpha| \cdot \| A \|$.
\end{itemize}
\end{fact}

\subsection{Basic Hyperbolic Functions: Scalar Version}
\label{sub:preli:scalar_version}

In this section, we analyze the properties of the hyperbolic functions, including $\sinh$ and $\cosh$, and exponential functions, $\exp$, where the elements of the domains of these functions are all scalars.

\begin{fact}\label{fac:exp_cosh_sinh_scalar} 
For a scalar $z \in \R$, we have
\begin{itemize}
    \item Part 1. $\cosh^2(z) - \sinh^2(z) = 1$
    \item Part 2. $|\exp(z)| \leq \exp(|z|)$
    \item Part 3. $|\cosh(z)| = \cosh(z) = \cosh(|z|) \leq \exp(|z|)$  
    \item Part 4. $|\sinh(z)| =  \sinh(|z|) \leq \exp(|z|)$
    \item Part 5. $\sinh(|z)| \leq \cosh(|z|) \leq \exp(|z|)$
\end{itemize}
Taylor expansions
\begin{itemize}
    \item $\exp(z) = \sum_{i=0}^{\infty} \frac{1}{i!} z^i$
    \item $\cosh(z) = \sum_{i=0}^{\infty} \frac{1}{(2i)!} z^{2i} $ 
    \item $\sinh(z) = \sum_{i=0}^{\infty} \frac{1}{(2i+1)!} z^{2i+1} $
\end{itemize}
Approximation in small range, 
\begin{itemize}
    \item For all $x \in \R$ satisfy that $ |x| \leq 0.1$, we can get $|\exp(x) - 1| \leq 2|x|$
    \item For all $x \in \R$ satisfy that $|x| \leq 0.1$, we can get $|\cosh(x) - 1| \leq x^2$
    \item For all $x \in \R$ satisfy that $|x| \leq 0.1$, we can get $|\sinh(x) | \leq 2|x|$
    \item For all $x,y\in \R$ satisfy that $| x - y | \leq 0.1$, we can get $| \exp(x) - \exp(y) | \leq \exp(x) \cdot 2 |x-y|$
    \item For all $x,y \in \R$ satisfy that $| x - y | \leq 0.1$, we can get $| \cosh(x) - \cosh(y) | \leq \cosh(x) \cdot 2|x - y|$
    \item For all $x,y\in \R$ satisfy that $| x - y | \leq 0.1$, we can get $| \sinh(x) - \sinh(y) | \leq \cosh(x) \cdot 2|x - y|$
\end{itemize}
\end{fact}
\begin{proof}
Most of the proofs are trivial or standard. We only provide some proofs.

{\bf Proof of Part 4.}

We have
\begin{align*}
| \sinh(z) | = & ~ | 0.5\exp(z) - 0.5\exp(-z) | \\
= & ~  0.5 \exp( |z| ) - 0.5 \exp(-|z|) \\
= & ~ \sinh(|z|)
\end{align*}
where second step is true because it's for $z\geq 0$ and also true for $z<0$.

We have
\begin{align*}
| \sinh(z) | = & ~ | 0.5 \exp(z) -0.5 \exp(-z)| \\
\leq & ~ 0.5 \exp(|z|) + 0.5 \exp(|z|) \\
= & ~ \exp(|z|)
\end{align*}

{\bf Proof of Part 5.}

We have
\begin{align*}
\sinh(|z|) = & ~ 0.5 \exp( |z| ) - 0.5 \exp(-|z|) \\
\leq & ~ 0.5 \exp( |z| ) + 0.5 \exp(-|z|) \\
= & ~ \cosh(|z)|
\end{align*}
We have
\begin{align*}
\cosh(|z)| = & ~ 0.5 \exp( |z| ) + 0.5 \exp(-|z|) \\
\leq & ~ 0.5 \exp(|z|) + 0.5 \exp(|z|) \\
= & ~ \exp(|z|)
\end{align*}

\end{proof}

\subsection{Basic Hyperbolic Functions: Vector Version}
\label{sub:preli:vector_version}

In this section, we keep analyzing the properties of the hyperbolic functions, namely $\sinh$ and $\cosh$, and exponential functions, $\exp$, but the elements of the domains of these functions are all vectors.

\begin{fact}[Formal version of Fact~\ref{fac:exp_cosh_sinh_vector_informal}]\label{fac:exp_cosh_sinh_vector}
For vectors $a, b \in \R^n$
\begin{itemize}
    \item $\| \exp(a) \|_{\infty} \leq \exp(\| a \|_2)$
    \item $\| \cosh(a) \|_{\infty} \leq \cosh( \| a \|_2 ) \leq \exp( \| a \|_2 )$
    \item $\| \sinh(a) \|_{\infty} \leq \sinh(\| a \|_2) \leq \cosh(\| a \|_2) \leq \exp(\| a \|_2)$
    \item $\cosh(a) \circ \cosh(a) - \sinh(a) \circ \sinh(a) ={\bf 1}_n$
\end{itemize}
Approximation in a small range,
\begin{itemize}
    \item For any $\| a - b \|_{\infty} \leq 0.01$, we have $\| \exp(a) - \exp(b) \|_2 \leq \| \exp(a) \|_2 \cdot 2 \| a - b \|_{\infty}$
    \item For any $\| a - b \|_{\infty} \leq 0.01$, we have $\| \cosh(a) - \cosh(b) \|_2 \leq \| \cosh(a) \|_2 \cdot 2 \| a - b \|_{\infty}$
    \item For any $\| a - b \|_{\infty} \leq 0.01$, we have $\| \sinh(a) - \sinh(b) \|_2 \leq \| \cosh(a) \|_2 \cdot 2 \| a - b \|_{\infty}$
\end{itemize}
\end{fact}

\begin{proof}
    Since $\exp, \cosh, \sinh$ are all applied entrywisely, it directly follows from Fact~\ref{fac:exp_cosh_sinh_scalar}.
\end{proof}

\subsection{Regularization}
\label{sub:preli:regularization}

Now, we define the regularization function, $L_{\reg}$, and analyze its properties.

\begin{definition}[Restatement of Definition~\ref{def:L_reg:informal}]\label{def:L_reg}
Given a matrix $A \in \R^{ n \times d}$ and $W = \diag (w) \in \R^{n \times n} $ where $w \in \R^n$ is a vector, we define $L_{\reg} : \R^d \rightarrow \R$
\begin{align*}
L_{\reg} (x) := 0.5 \| W A x\|_2^2    
\end{align*} 
\end{definition}

\begin{lemma}[Folklore, see \cite{lsz23,dls23} as an example]\label{lem:hessian_L_reg}
If the following condition holds 
\begin{itemize}
    \item Let $L_{\reg}(x)$ be defined as Definition~\ref{def:L_reg}.
\end{itemize}
Then, 
we have 
\begin{itemize}
    \item The gradient is
    \begin{align*}
        \frac{ \d L_{\reg}} { \d x} = A^\top W^2  A x
    \end{align*}
    \item The Hessian is 
    \begin{align*}
        \frac{\d^2 L_{\reg}}{ \d x^2} = A^\top W^2 A    
    \end{align*}
\end{itemize}
\end{lemma}

\subsection{Gradient and Hessian}
\label{sub:preli:gradient}

Finally, in this section, we define the gradient and Hessian of the loss function and present the definition of the update of the Newton method.

\begin{definition}[Gradient and Hessian]
Let $L(x)$ be some loss function. 
The gradient $g : \R^d \rightarrow \R^d$ of the loss function is defined as 
\begin{align*}
    g(x) := \nabla L(x) = \frac{\d L}{\d x}
\end{align*}

The Hessian $H : \R^d \rightarrow \R^{d \times d}$ of the loss function is defined as follow:
\begin{align*}
    H(x) := \nabla^2 L(x) = \frac{\d^2 L }{\d x^2}
\end{align*}

\end{definition}

\begin{definition}[Update of the Newton method]\label{def:exact_update_variant}
Given that the gradient function $g: \R^d \rightarrow \R^d$ and the Hessian matrix $H : \R^d \rightarrow \R^{d \times d}$, the exact process of the Newton method can be defined as follows: 
\begin{align*}
    x_{t+1} = x_t - H(x_t)^{-1} \cdot g(x_t)
\end{align*}

\end{definition}

%% file: g_H.tex
\section{General Function: Gradient and Hessian Computations}\label{sec:g_H}

In Section~\ref{sub:g_H:gradient}, we compute the gradients of $u(x)$, $\alpha(x)$, $c(x)$, and $L_u(x)$. In Section~\ref{sub:g_H:step_1}, we present the second-order derivatives of $u(x)$ with respect to $x_i^2$ and $x_ix_j$, where $x_i$ and $x_j$ are two arbitrary entries of the vector $x \in \R^d$. In Section~\ref{sub:g_H:step_2}, we present the second-order derivatives of $\alpha(x)$ with respect to $x_i^2$ and $x_ix_j$, where $x_i$ and $x_j$ are two arbitrary entries of the vector $x \in \R^d$. In Section~\ref{sub:g_H:step_3}, we compute the second-order derivatives of $c(x)$ with respect to $x_i^2$ and $x_ix_j$. Finally, in Section~\ref{sub:g_H:step_4}, we compute the second-order derivatives of $L_u(x)$ with respect to $x_i^2$ and $x_ix_j$.

\subsection{Gradient Computations}
\label{sub:g_H:gradient}

In this section, we compute the gradients of $u(x)$, $\alpha(x)$, $c(x)$, and $L_u(x)$, namely their first-order derivative with respect to $x_i$.

\begin{lemma}[Gradient]\label{lem:gradient_computation}
If the following conditions hold
\begin{itemize}
    \item Let $A \in \R^{n \times d}$ and $b \in \R^n$.
    \item For all $i \in [d]$, $A_{*,i} \in \R^n$ denotes the $i$-th column of matrix $A \in \R^{n \times d}$.
    \item Let $u(x)$ be defined in Definition~\ref{def:u}.
    \item Let $v(x)$ be defined in Definition~\ref{def:v}.
    \item Let $\alpha(x)$ be defined in Definition~\ref{def:alpha}.
    \item Let $c(x)$ be defined in Definition~\ref{def:c}.
    \item Let $L_{u}(x)$ be defined in Definition~\ref{def:L_u} 
\end{itemize}
Then, for each $i \in [d]$, we have
\begin{itemize}
    \item Part 1. (see Part 1 in Lemma 5.6 in page 11 in \cite{dls23})  
    \begin{align*}
        \frac{ \d u(x) }{ \d x_i} =v(x) \circ A_{*,i}
    \end{align*}
    \item Part 2. (see Part 2 in Lemma 5.6 in page 11 in \cite{dls23})
    \begin{align*}
        \frac{\d \alpha(x) }{\d x_i} = \langle v(x), A_{*,i}\rangle
    \end{align*}
    \item Part 3. 
    \begin{align*}
        \frac{\d c(x)}{ \d x_i} = v(x) \circ A_{*,i} - b \cdot \langle v(x), A_{*,i} \rangle
    \end{align*} 
    \item Part 4.  
    \begin{align*}
        \frac{\d L_{u} (x) }{\d x_i} = A_{*,i}^\top (  c(x) \circ v(x) - v(x) \langle b, c(x) \rangle )  
    \end{align*}
\end{itemize}
\end{lemma}
\begin{proof}
    {\bf Proof of Part 1.}
    For each $j \in [n]$, we have 
    \begin{align*}
     \frac{ \d ( u(x) )_j }{ \d x_i } 
     = & ~ v(x)_j \cdot \frac{\d (Ax)_j}{\d x_i} \\
     = & ~ v(x)_j \cdot \frac{ (A \d x)_j}{\d x_i} \\
     = & ~ v(x)_j \cdot {A_{j,i}}
    \end{align*}
    where the first step follows from chain rule (Fact~\ref{fac:chain_rule}), the second step follows from basic chain rule (Fact~\ref{fac:chain_rule}), the third step follows from basic calculus rule (Fact~\ref{fac:chain_rule}). 
    
    Since the above equation is true for all $j \in [n]$, we have 
    \begin{align*}
        \underbrace{ \frac{ \d u(x) }{ \d x_i} }_{n \times 1} = \begin{bmatrix} 
        \frac{\d u(x)_1}{\d x_i} \\
        \frac{\d u(x)_2}{\d x_i} \\
        \vdots \\
        \frac{\d u(x)_n}{\d x_i} 
        \end{bmatrix} 
        = \underbrace{ v(x) }_{n \times 1} \circ \underbrace{ A_{*,i} }_{n \times 1}
    \end{align*}
    
    {\bf Proof of Part 2.}
    It trivially follows from arguments in {\bf Part 1}. 

    {\bf Proof of Part 3.}

    \begin{align*}
        \frac{\d c(x)}{ \d x_i} 
        = & ~ \frac{\d }{\d x_i} ( u(x) - b \cdot \alpha(x) ) \\
        = & ~ v(x) \circ A_{*,i} - b \cdot \langle v(x), A_{*,i} \rangle
    \end{align*}
    where the first step is due to the definition of $c(x)$ (see Definition~\ref{def:c}), the second step follows from {\bf Part 1} and {\bf Part 2}.

    {\bf Proof of Part 4.}
    \begin{align*}
        \frac{\d L_{u}(x)}{\d x_i}
        = & ~ \frac{\d}{\d x_i}(0.5 \cdot \| c(x) \|_2^2) \\
        = & ~ c(x)^\top \frac{\d}{\d x_i} c(x) \\
        = & ~ c(x)^\top ( v(x) \circ A_{*,i} - b \cdot \langle v(x), A_{*,i} \rangle ) \\
        = & ~ A_{*,i}^\top ( c(x) \circ v(x) ) - A_{*,i}^\top v(x) \langle b, c(x) \rangle \\
        = & ~ A_{*,i}^\top (  c(x) \circ v(x) - v(x) \langle b, c(x) \rangle )  
    \end{align*}
where the first step is due to the definition of $L_{u}(x)$ (see Definition~\ref{def:L_u}), the second step follows from chain rule (Fact~\ref{fac:chain_rule}), the third step is due to {\bf Part 3}, the fourth step is obtained by using Fact~\ref{fac:circ_diag}, 
and the fifth step follows from simple algebra.
\end{proof}

\subsection{Hessian Computation Step 1.  Vector Function \texorpdfstring{$\exp(Ax)$}{}}
\label{sub:g_H:step_1}

We state a tool from previous work \citep{lsz23,dls23}.
\begin{lemma}[Lemma 5.9 in \cite{dls23} and implicitly in \cite{lsz23}]\label{lem:hessian_u}
If the following conditions hold
\begin{itemize}
    \item Let $A \in \R^{n \times d}$
    \item Let $x \in \R^d$.
\end{itemize}
We have
\begin{itemize}
    \item Part 1.
    \begin{align*}
        \frac{\d^2 u(x) }{ \d x_i^2} = A_{*,i} \circ u(x) \circ A_{*,i}
    \end{align*}
    \item Part 2.
    \begin{align*}
        \frac{\d^2 u(x) }{ \d x_i \d x_j} = A_{*,j} \circ u(x) \circ A_{*,i}.
    \end{align*}
\end{itemize}
\end{lemma}

\subsection{Hessian Computation Step 2. Scalar Function \texorpdfstring{$\alpha(x)$}{}}
\label{sub:g_H:step_2}

We state a tool from previous work \citep{lsz23,dls23}.
\begin{lemma}[Lemma 5.10 in \cite{dls23}, implicitly in \cite{lsz23}]\label{lem:hessian_alpha}
If the following conditions hold 
\begin{itemize}
    \item Let $\alpha(x)$ be defined as
    Definition~\ref{def:alpha}.
    \item Let $u(x)$ be defined as in Definition~\ref{def:u}.
    \item Let $A \in \R^{n \times d}$.
    \item Let $x \in \R^d$.
\end{itemize}
Then, we have
\begin{itemize}
\item Part 1.
    \begin{align*}
        \frac{\d^2 \alpha(x) }{\d x_i^2} = \langle u(x), A_{*,i} \circ A_{*,i} \rangle
    \end{align*}
    \item Part 2.
    \begin{align*}
        \frac{\d^2 \alpha(x) }{\d x_i \d x_j} = \langle u(x),  A_{*,i} \circ A_{*,j} \rangle
    \end{align*}
\end{itemize}
\end{lemma}

\subsection{Hessian Computation Step 3. Vector Function \texorpdfstring{$c(x)$}{}}
\label{sub:g_H:step_3}

Now, we compute the second-order derivative of $c(x)$ with respect to $x_i^2$ and $x_ix_j$.

\begin{lemma} \label{lem:hessian_c}
If the following conditions hold  
\begin{itemize}
    \item Let $c(x)$ be defined as Definition~\ref{def:c}.
    \item Let $A \in \R^{n \times d}$.
    \item Let $x \in \R^d$.
    \item Let $b \in \R^n$.
\end{itemize}
Then, we have
\begin{itemize}
\item Part 1.
    \begin{align*}
\frac{\d^2 c(x) }{\d x_i^2}=A_{*,i} \circ u(x) \circ A_{*,i}  - b \cdot \langle u(x), A_{*,i} \circ A_{*,i} \rangle
    \end{align*}
    \item Part 2.
    \begin{align*}
       \frac{\d^2 c(x) }{\d x_i \d x_j} = A_{*,i} \circ u(x) \circ A_{*,j}- b \cdot \langle u(x), A_{*,i} \circ A_{*,j} \rangle 
    \end{align*}
\end{itemize}
\end{lemma}
\begin{proof}

{\bf Proof of Part 1.}

\begin{align*}
\frac{\d^2 c(x) }{\d x_i^2} 
= & ~ \frac{\d^2 }{\d x_i^2} ( u(x) - b \cdot \alpha(x) ) \\
= & ~ A_{*,i} \circ u(x) \circ A_{*,i} - b \cdot \langle u(x), A_{*,i} \circ A_{*,i} \rangle 
\end{align*}
where the first step follows from  definition of $c(x)$ (see Definition~\ref{def:c}), the second step follows from Lemma~\ref{lem:hessian_u} and Lemma~\ref{lem:hessian_alpha}.

{\bf Proof of Part 2.}

\begin{align*}
  \frac{\d^2 c(x) }{\d x_i \d x_j}
= & ~  \frac{\d^2  }{ \d x_i \d x_j } ( u(x) - b \cdot \alpha(x) ) \\
= & ~ A_{*,i} \circ u(x) \circ A_{*,j} - b \cdot \langle u(x) , A_{*,i} \circ A_{*,j} \rangle 
\end{align*}
where the first step follows from definition of $c(x)$ (see Definition~\ref{def:c}), the second step follows from Lemma~\ref{lem:hessian_u} and Lemma~\ref{lem:hessian_alpha}.
\end{proof}

\subsection{Hessian Computation Step 4. Scalar Function \texorpdfstring{$L_{u}(x)$}{}}
\label{sub:g_H:step_4}

Then, we compute the second-order derivative of $L_u(x)$ with respect to $x_i^2$ and $x_ix_j$, by first introducing some functions, $B_{1, 1}, B_{1, 2}, B_{1, 3}, B_{1, 4}, B_{2, 1}, B_{2, 2}$ (see Definition~\ref{def:B}), to simplify the process of computation.

\begin{definition}\label{def:B}
Given the following objects
\begin{itemize}
\item Let $A \in \R^{n \times d}$.
\item Let $x \in \R^d$.
\item Let $b \in \R^n$.
\end{itemize}

Then, we define the functions $B_{1, 1}, B_{1, 2}, B_{1, 3}, B_{1, 4}, B_{2, 1}, B_{2, 2} : \R^d \rightarrow \R^{n \times n}$ as
\begin{align*}
    B_{1,1}(x) := & ~ \diag( v(x) \circ v(x) )  \\
    B_{1,2}(x) := & ~ -( v(x) \circ b ) \cdot v(x)^\top \\
    B_{1,3}(x) := & ~ - v(x) \cdot ( v(x) \circ b)^\top \\
    B_{1,4}(x) := & ~ \| b \|_2^2 \cdot v(x) v(x)^\top 
\end{align*}
We define
\begin{align*}
    B_{2,1}(x) := & ~ \diag(c(x) \circ u(x) ) \\
    B_{2,2}(x) := & ~ -\langle c(x),b \rangle \diag( u(x) )
\end{align*}

    We define $B : \R^d \rightarrow \R^{n \times n}$
    as follows:
\begin{align*}
    B(x):= & ~ B_{1,1}(x) + B_{1,2}(x) + B_{1,3}(x) + B_{1,4}(x) \\
    & ~ + B_{2,1}(x) + B_{2,2}(x)
\end{align*}
\end{definition}

\begin{lemma}\label{lem:hessian_L_u}
If the following conditions hold
\begin{itemize}
    \item Let $B(x)$ be defined as in Definition~\ref{def:B}.
    \item Let $A \in \R^{n \times d}$.
    \item Let $L_{u}(x)$ be defined as in Definition~\ref{def:L_u}.
\end{itemize}
Then, we have
\begin{itemize}
\item Part 1.
    \begin{align*}
        \frac{\d^2 L_{u}(x) }{\d x_i^2} = A_{*,i}^\top B A_{*,i}
    \end{align*}
    \item Part 2.
    \begin{align*}
        \frac{\d^2 L_{u}(x) }{\d x_i \d x_j} = A_{*,i}^\top B A_{*,j}
    \end{align*}
\end{itemize}
\end{lemma}

\begin{proof}

{\bf Proof of Part 1.}

\begin{align*}
\frac{\d^2 L_{u}(x)}{\d x_i^2}
= & ~ \frac{ \d }{ \d x_i } ( \frac{ \d L_{u}(x) }{ \d x_i }) \\
= & ~ \frac{ \d }{ \d x_i } (c(x)^\top \frac{\d c(x)}{\d x_i})\\
= & ~ \langle \frac{\d c(x)}{\d x_i}, \frac{ \d c(x)}{\d x_i} \rangle + \langle c(x), \frac{\d^2 c(x)}{\d x_i^2 } \rangle
\end{align*}
where the first step follows from simple algebra, the second step follows from basic chain rule (see Fact~\ref{fac:chain_rule}), and the last step follows from basic calculus.

For the first term, in the above equation, we have
\begin{align*}
\langle \frac{\d c(x)}{\d x_i}, \frac{ \d c(x)}{\d x_i} \rangle
= & ~ \| v(x) \circ A_{*,i} - b \cdot \langle  v(x), A_{*,i} \rangle\|_2^2 \\
= & ~ A_{*,i}^\top \diag( v(x) \circ v(x) ) A_{*,i} \\
& ~ - A_{*,i}^\top ( v(x) \circ b ) v(x)^\top A_{*,i} \\
& ~ - A_{*,i}^\top ( v(x) ) ( v(x) \circ b)^\top A_{*,i} \\
& ~ + A_{*,i}^\top \| b \|_2^2 v(x) v(x)^\top A_{*,i} \\
= & ~ A_{*,i}^\top ( B_{1,1}(x) + B_{1,2}(x) + B_{1,3}(x) + B_{1,4}(x) ) A_{*,i}
\end{align*}
where the first step is due to {\bf Part 3} of Lemma~\ref{lem:gradient_computation}, the second step follows from Fact~\ref{fac:inner_product}, and the last step follows from the definition of $B_{1,i}(x)$ for each $i \in [4]$ (see Definition~\ref{def:B}).

For the second term, we have
\begin{align*}
\langle c(x), \frac{\d^2 c(x)}{\d x_i^2 } \rangle 
= & ~ \langle c(x), A_{*,i} \circ u(x) \circ A_{*,i}- b \cdot \langle u(x),  A_{*,i} \circ A_{*,i} \rangle \rangle  \\
= & ~ A_{*,i}^\top \diag(c(x) \circ u(x) ) A_{*,i} \\
& ~ -  A_{*,i}^\top \langle c(x),b \rangle \diag( u(x) ) A_{*,i} \\
= & ~ A_{*,i}^\top ( B_{2,1}(x) + B_{2,2}(x)) A_{*,i} 
\end{align*}
where the first step is due to {\bf Part 1} of  Lemma~\ref{lem:hessian_c}, the second step follows from Fact~\ref{fac:inner_product}, and the last step follows from $B_{2,i}$ for all $i \in [2]$ (see Definition~\ref{def:B}).

Thus, we finally have
\begin{align*}
\frac{\d^2 L_{u}(x) }{ \d x_i^2} 
=  A_{*,i}^\top B(x) A_{*,i}
\end{align*}

{\bf Proof of Part 2.}

The proof is similar, and we omitted the details here.
\end{proof}

%% file: pd.tex
\section{General Function: Psd Lower Bound}
\label{sec:PSD}

In Section~\ref{sub:PSD:upper_bound}, we provide the upper bound for the $\ell_2$ norms of $u(x), v(x), c(x) \in \R^n$ and for the absolute value of $\alpha(x) \in \R$. In Section~\ref{sub:PSD:bound}, we compute both the upper bound and the lower bound of $B(x)$ in terms of $\preceq$. In Section~\ref{sub:PSD:lower_bound}, we analyze the lower bound of Hessian.

\subsection{Upper Bound for Several Basic Quantities}
\label{sub:PSD:upper_bound}

In this section, we compute the bounds for the $\ell_2$ norms of the vectors $u(x), v(x), c(x) \in \R^n$ and compute the bound for the absolute value of $\alpha(x)$.

\begin{claim}\label{cla:bound_u_alpha_c}
If the following conditions hold
\begin{itemize}
    \item Let $R \geq 2$.
    \item $\| A \| \leq R$
    \item $\| x \|_2 \leq R$
    \item $\| b \|_2 \leq R$
    \item Let $u(x) \in \R^n$ be defined as Definition~\ref{def:u}.
    \item Let $v(x) \in \R^n$ be defined as Definition~\ref{def:v}.
    \item Let $\alpha(x) \in \R$ be defined as  Definition~\ref{def:alpha}.
    \item Let $c(x) \in \R^n$ be defined as in Definition~\ref{def:c}. 
\end{itemize}
Then, we have
\begin{itemize}
\item Part 1. (see \cite{lsz23,dls23,lsx+23})
\begin{align*}
    \| u(x) \|_2 \leq & ~ \sqrt{n} \exp(R^2) \\
    \| v(x) \|_2 \leq & ~ \sqrt{n} \exp(R^2) 
\end{align*}
\item Part 2.
\begin{align*}
    |\alpha(x)| \leq n \exp(R^2)
\end{align*}
\item Part 3. 
\begin{align*}
    \| c (x) \|_2 \leq n R \exp(R^2)
\end{align*}
\end{itemize} 
\end{claim}
\begin{proof}

{\bf Proof of Part 1.}
The proof is standard in the literature, and we omit the details here.

{\bf Proof of Part 2.}
We can show  
\begin{align*}
|\alpha(x)| = & ~ | \langle u(x), {\bf 1}_n \rangle | \\
\leq & ~ \sqrt{n} \cdot \| u(x) \|_2   \\
\leq & ~ n \cdot \exp(R^2)
\end{align*}
where the first step follows from definition of $\alpha(x)$ (see Definition~\ref{def:alpha}), the second step follows from Fact~\ref{fac:basic_vector_norm}, the third step follows from {\bf Part 1}.

{\bf Proof of Part 3.}
We can show
\begin{align*}
\| c(x) \|_2  
= & ~ \| u(x) - \alpha(x) b \|_2 \\
\leq & ~ \| u(x) \|_2 + \| \alpha(x) b \|_2 \\
= & ~ \sqrt{n} \exp(R^2) + | \alpha(x) | \cdot \| b \|_2 \\
\leq & ~ \sqrt{n} \exp(R^2) + | \alpha(x) | \cdot R \\
\leq & ~ \sqrt{n} \exp(R^2) + n R \exp(R^2) \\
\leq & ~ 2 n R \exp(R^2),
\end{align*}
where the first step comes from the definition of $c(x)$ (see Definition~\ref{def:c}), the second step follows from the triangle inequality, the third step is because of {\bf Part 1}, the fourth step follows from the assumption on $b$, the fifth step follows from {\bf Part 2}, and the last step follows from simple algebra.

\end{proof}

\subsection{PSD Bounds for Several Basic Matrix Functions}
\label{sub:PSD:bound}

In this section, we first define the matrices $B_{\rank}^1,  B_{\rank}^2,  B_{\rank}^3, B_{\diag}^1,  B_{\diag}^2 \in \R^{n \times n}$ and find the $\preceq$-bound for them. Then, we combine them together to form the bound for $B(x) \in \R^{n \times n}$

\begin{definition}\label{def:B_rank_diag}

Given the following objects
\begin{itemize}
\item Let $u(x)$ be defined as in Definition~\ref{def:u}.
\item Let $c(x)$ be defined as in Definition~\ref{def:c}.
\item Let $b \in \R^n$.
\end{itemize}

    We define $B : \R^d \rightarrow \R^{n \times n}$ as follows:
\begin{align*}
    B(x):= B_{\rank} + B_{\diag}
\end{align*}
We define
\begin{align*}
    B_{\rank} := & ~  B_{\rank}^1+  B_{\rank}^2 +  B_{\rank}^3 \\
    B_{\diag} := & ~  B_{\diag}^1+  B_{\diag}^2 
\end{align*}
We define
\begin{itemize}
    \item $B_{\rank}^1 := -u(x)( u(x) \circ b ) ^\top $
    \item $B_{\rank}^2 := -( u(x) \circ b ) u(x)^\top$
    \item $B_{\rank}^3 := \| b \|_2^2 u(x) u(x)^\top$
    \item $B_{\diag}^1 := \diag( ( u(x) + c(x) ) \circ u(x) + q)$
    \begin{itemize}
        \item $q = {\bf 0}_n$ (when $u(x) = \exp(Ax)$)
        \item $q = -{\bf 1}_n$ (when $u(x) = \cosh(Ax)$)
        \item $q = {\bf 1}_n$ (when $u(x) = \sinh(Ax)$)
    \end{itemize}
    \item $B_{\diag}^2 := -\langle c(x),b \rangle \diag(u(x))$
\end{itemize}

\end{definition}

\begin{lemma}\label{lem:B_psd_bound}
If the following situations hold
\begin{itemize}
\item $B(x)$ is a $n \times n$ dimension matrix (See Definition~\ref{def:B_rank_diag}).
\item $B_{\rank}^1$, $B_{\rank}^2$, $B_{\rank}^3$ are defined in Definition~\ref{def:B_rank_diag}. 
\item $B_{\diag}^1$, $B_{\diag}^2$ are defined in Definition~\ref{def:B_rank_diag}. 
\end{itemize}
Then, we have
\begin{itemize}
    \item Part 1.
    \begin{align*}
        -\| b \|_2 v(x) v(x)^\top \preceq  B_{\rank}^1  \preceq \| b \|_2 v(x) v(x)^\top
    \end{align*}
    \item Part 2.
    \begin{align*}
         -\|b\|_2 v(x) v(x)^\top \preceq B_{\rank}^2 \preceq \|b\|_2 v(x) v(x)^\top
    \end{align*}
    \item Part 3.
    \begin{align*}
        B_{\rank}^3 = \| b \|_2^2 v(x) v(x)^\top
    \end{align*}
    \item Part 4.
    \begin{align*}
         - ( 1 + (  \| u(x) \|_{\infty} + \| c(x) \|_{\infty} )  \cdot \| u(x) \|_{\infty} ) \cdot I_n \preceq B_{\diag}^1 \preceq (1+ (  \| u (x) \|_{\infty} +  \| c(x) \|_{\infty} ) \cdot \| u(x) \|_{\infty} ) \cdot I_n
    \end{align*}
    \item Part 5.
    \begin{align*}
        - \| b \|_2 \| c(x) \|_2 \| u(x) \|_{\infty} I_n \preceq B_{\diag}^2 \preceq \| b \|_2 \| c(x) \|_2 \| u(x) \|_{\infty} I_n
    \end{align*}
    \item Part 6.
    \begin{itemize}
        \item Let $R_0 = \max\{ \| u (x) \|_2, \| v(x) \|_2,  \| b \|_2, \| c(x) \|_2, 1\}$
        \item Then, we have
        \begin{align*}
            -10 R_0^4 \cdot I_n \preceq B(x) \preceq 10 R_0^4 \cdot I_n
        \end{align*}
    \end{itemize}
\end{itemize}
\end{lemma}
\begin{proof}

{\bf Proof of Part 1.}
First, we focus on the lower bound of $B_{\rank}^1$. We have
\begin{align*}
    B_{\rank}^1 
    = & ~ -v(x)( v(x) \circ b ) ^\top \\
    \succeq & ~ - \| b \|_2 \cdot v(x) v(x)^\top,
\end{align*}
where the first step follows from the definition of $B_{\rank}^1$ (see Definition~\ref{def:B_rank_diag}) and the second step follows from Fact~\ref{fac:psd}. 

Similarly, we have
\begin{align*}
 B_{\rank}^1 = & ~ -v(x)( v(x) \circ b ) ^\top \\
    \preceq & ~ \|b\|_2 \cdot v(x)  v(x)^\top,
\end{align*}
where the first step follows from the definition of $B_{\rank}^1$ (see Definition~\ref{def:B_rank_diag}) and the second step follows from Fact~\ref{fac:psd} 

{ \bf Proof of Part 2.}
According to what we obtained in the Part 1, we can also have
\begin{align*}
     -\|b\|_2 v(x) v(x)^\top \preceq B_{\rank}^2 \preceq \|b\|_2 v(x) v(x)^\top
\end{align*}

{ \bf Proof of Part 3.}

The proof is trivially following from definition of $B_{\rank}^3$. 
We  have  
\begin{align*}
    B_{\rank}^3 = \| b \|_2^2 \cdot v(x) v(x)^\top
\end{align*}

{ \bf Proof of Part 4.}
For $i\in [n]$, $u(x)_i>0$, we have 
\begin{align*}
     B_{\diag}^1 
     = & ~ \diag( ( u(x) + c(x) ) \circ u(x) + q)\\
    \preceq & ~ ( 1+ ( \| u(x) \|_{\infty} +   \| c(x) \|_{\infty} ) \| u(x) \|_{\infty} ) \cdot I_n,
\end{align*}
where the first step is due to the definition of $B_{\diag}^1$ (see Definition~\ref{def:B_rank_diag}) and the second step follows from Fact~\ref{fac:psd}.  

On the other hand, we have
\begin{align*}
B_{\diag}^1 \succeq - (1 + ( \| u(x) \|_{\infty} + \| c(x) \|_{\infty} ) \| u(x) \|_{\infty} ) \cdot I_n
\end{align*}

{\bf Proof of Part 5.}

\begin{align*}
    B_{\diag}^2 
    = & ~ - \langle c(x), b \rangle \diag(u(x) ) \\
    \preceq  & ~  \| b \|_2 \cdot \| c(x) \|_2 \cdot \diag(u(x) ) \\
    \preceq  & ~  \| b \|_2 \cdot \| c(x) \|_2 \cdot \| u(x) \|_{\infty} \cdot I_n,
\end{align*}
where the first step follows from the definition of $B_{\diag}^2$ (see Definition~\ref{def:B_rank_diag}), the second step follows from Fact~\ref{fac:psd}, and the third step follows from Fact~\ref{fac:psd}.

Similarly, we have
\begin{align*}
     B_{\diag}^2 
     = & ~ - \langle c(x), b \rangle \diag(u(x) ) \\
     \succeq & ~ -\| b \|_2 \cdot \| c(x) \|_2 \cdot \diag(u(x) )\\
    \succeq  & ~ -\| b \|_2 \cdot \| c(x) \|_2 \cdot \| u(x) \|_{\infty} \cdot I_n,
\end{align*}
where the first step comes from the definition of $B_{\diag}^2$ (see Definition~\ref{def:B_rank_diag}), the second step follows from Fact~\ref{fac:psd}, and the third step follows from Fact~\ref{fac:psd}.

{ \bf Proof of Part 6.}
Using Fact~\ref{fac:psd}
\begin{align*}
u(x) u(x)^\top \preceq \| u (x) \|_2^2 I_n
\end{align*}

We also have
\begin{align*}
    \max \{ B^1_{\rank},  B^2_{\rank}, B^3_{\rank}, B^1_{\diag}, B^2_{\diag}\} \leq 2R_0^4 \cdot I_n
\end{align*}

\end{proof}

\subsection{Lower bound on Hessian}
\label{sub:PSD:lower_bound}

In this section, we compute the lower bound for Hessian.

\begin{lemma} \label{lem:hessian_psd_exp}
If conditions as follows are satisfied
\begin{itemize}
    \item Let $A \in \R^{n \times d}$.
    \item Let $u(x)$ be defined as Definition~\ref{def:u}.
    \item Let $v(x)$ be defined as Definition~\ref{def:v}.
    \item $L_{u}(x)$ is defined in Definition~\ref{def:L_u}.
    \item $L_{\reg}(x)$ is defined in Definition~\ref{def:L_reg}.
    \item $L(x)=L_{\reg}(x)+L_{u}(x)$.
    \item Given $w\in \R^n$, $W = \diag(w) \in \R^{n \times n}$ and $W^2$ denotes the matrix with $w_i^2$ as the $i$-th diagonal. 
    \item We use $\sigma_{\min} (A)$ as the minimum singular value of $A$.
    \item We let $l>0$ as a scalar. 
    \item Let $R_0 = \max\{ \| u (x) \|_2, \| v \|_2, \| b \|_2, \| c(x) \|_2, 1\}$.
\end{itemize}
Then we have
\begin{itemize}
    \item Part~1. If all $i\in [n]$, $w_i^2 \geq 10 R_0^4 + l/\sigma_{\min} (A)^2$, then we have 
    \begin{align*}
        \frac{\d^2 L}{ \d x^2} \succeq l \cdot I_d
    \end{align*}
    \item Part~2. If all $i\in [n]$, $w_i^2 \geq 200 R_0^4 + l/\sigma_{\min} (A)^2$, then we have
    \begin{align*}
        (1-1/10)\cdot (B(x)+W^2)\preceq W^2 \preceq (1+1/10)\cdot (B(x)+W^2).
    \end{align*}
\end{itemize}
\end{lemma}

\begin{proof}
By Lemma~\ref{lem:hessian_L_u}, we have
\begin{align}\label{eq:hessian_L_u}
     \frac{\d^2 L_{u}}{\d x^2} = A^{\top} B(x) A,
\end{align}

By Lemma~\ref{lem:hessian_L_reg}, we have
\begin{align}\label{eq:hessian_L_reg}
    \frac{\d^2 L_{\reg}}{\d x^2} = A^\top W^2 A.
\end{align}

By what we have in the Lemma statement, we also have
\begin{align}\label{eq:lemma_statement}
     \frac{\d^2 L}{\d x^2} =  \frac{\d^2 L_{u}}{\d x^2} + \frac{\d^2 L_{\reg}}{\d x^2}
\end{align}

By combining Eq.~\eqref{eq:hessian_L_u}, Eq.~\eqref{eq:hessian_L_reg}, and Eq.~\eqref{eq:lemma_statement}, we can rewrite the equation above as follows:
\begin{align*}
    \frac{\d^2 L}{\d x^2} 
    = & ~ A^\top B(x) A + A^\top W^2 A \\
    = & ~ A^\top (B(x) + W^2 ) A,
\end{align*}
where the second step follows from simple algebra.

Now we define
\begin{align*}
    D : = B(x) + W^2 
\end{align*}
Now we get the bound of $D$
\begin{align*}
    D \succeq & ~ - 10 R_0^4  I_n + w_{\min}^2 I_n\\
      = & ~  (w^2_{\min}- 10 R_0^4) I_n\\
    \succeq & ~  \frac{ l }{ \sigma_{\min} (A)^2 } I_n,
\end{align*}
where the first step follows from {\bf Part 6} of Lemma~\ref{lem:B_psd_bound}, the second step follows from simple algebra, and the third step is because of the assumption of this part.

Since $D$ is positive definite, then we have
\begin{align*}
    A^\top D A \succeq \sigma_{\min} (D) \cdot \sigma_{\min} (A)^2 \cdot I_d \succeq l \cdot I_d
\end{align*}

{\bf Proof of Part 2.}

Using {\bf Part 6} of Lemma~\ref{lem:B_psd_bound}, we have
\begin{align*}
    -10 R_0^4 I_n \preceq B(x) \preceq 10 R_0^4 I_n.
\end{align*}

From assumption on $W$, we also have
\begin{align*}
W^2
\succeq & ~  200 R_0^4 I_n \\
- W^2 \preceq & ~ - 200 R_0^4 I_n
\end{align*}

Combining the above three equations,
\begin{align*}
-\frac{1}{20} W^2 \preceq B(x) \preceq \frac{1}{20} W^2
\end{align*}

Thus,
\begin{align*}
(1-\frac{1}{20} ) W^2 \preceq B(x) + W^2 \preceq (1+\frac{1}{20}) W^2
\end{align*}
which implies that
\begin{align*}
-  (1+\frac{1}{10}) (B(x) + W^2) \preceq W^2 \preceq (1+\frac{1}{10}) (B(x) + W^2)
\end{align*}

\end{proof}

%% file: lip.tex
\section{General Function: Hessian Is Lipschitz with Respect To \texorpdfstring{$x$}{}}
\label{sec:lipschitz}

In Section~\ref{sub:lipschitz:main}, we summarize all of the important properties that we derive in the following subsections to form an upper bound for $\|H(x) - H(y)\|$.
In Section~\ref{sub:lipschitz:ux}, we analyze the upper bound for $\|u(x) - u(y)\|_2$.
In Section~\ref{sub:lipschitz:alphax}, we analyze the upper bound for $|\alpha(x) - \alpha(y)|$.
In Section~\ref{sub:lipschitz:cx}, we prove the upper bound for $\|c(x) - c(y)\|_2$.
In Section~\ref{sub:lipschitz:summary}, we evaluate the upper bound of the sum of all the spectral norms of the matrices $G_i \in \R^{n \times n}$, for all $i \in [5]$, where the spectral norms of each of the matrix $G_i$ is evaluated in each of the following subsection.
In Section~\ref{sub:lipschitz:step1}, we analyze the upper bound of the spectral norm of $G_1 \in \R^{n \times n}$.
In Section~\ref{sub:lipschitz:step2}, we find the upper bound of the spectral norm of $G_2 \in \R^{n \times n}$.
In Section~\ref{sub:lipschitz:step3}, we study the upper bound of the spectral norm of $G_3 \in \R^{n \times n}$.
In Section~\ref{sub:lipschitz:step4}, we prove the upper bound of the spectral norm of $G_4 \in \R^{n \times n}$.
In Section~\ref{sub:lipschitz:step5}, we show the upper bound of the spectral norm of $G_5 \in \R^{n \times n}$.

\subsection{Main Result}
\label{sub:lipschitz:main}

In this section, we introduce our main result, which is the combination of all the important concepts in Section~\ref{sec:lipschitz}.

\begin{lemma}\label{lem:hessian_lipschitz}
If the following condition holds
\begin{itemize}
    \item Let $H(x) = \frac{ \d^2 L}{\d x^2} $  
    \item Let $R > 4$
    \item $\| x \|_2 \leq R$, $\| y \|_2 \leq R$, where $x, y \in \R^d$
    \item $\| A (x-y) \|_\infty < 0.01$, where $A \in \R^{n \times d}$
    \item $\| A \| \leq R$
    \item $\| b \|_2 \leq R$, where $b \in \R^n$
    \item Let $R_{\infty}:=\max\{ \|u(x) \|_2, \| u(y) \|_2, \| c(x) \|_2, \| c(y) \|_2, 1 \}$
    \begin{itemize}
    \item where $R_{\infty} \leq 2n R \exp(R^2)$ 
    \item this is proved by Part 1 and Part 3 in Claim~\ref{cla:bound_u_alpha_c}  
    \end{itemize}
\end{itemize}
Then we have 
\begin{align*}
    \| H(x) - H(y)\| \leq  n^{4} \exp (20 R^2) \cdot \| x - y \|_2
\end{align*}
\end{lemma}

\begin{proof}
\begin{align*}
    & ~ \| H(x) - H(y) \| \\
    \leq & ~ \| A \| \cdot (  \| G_1 \| + \|G_2\| + \cdots + \|G_5\|) \| A \| \\
    \leq & ~ R^2 \cdot (  \| G_1 \| + \| G_2 \| + \cdots + \| G_5 \|) \\
    \leq & ~ R^2 \cdot 5 \cdot R_{\infty}^3 \| b \|_2 \sqrt{n} \cdot \| u(x) - u(y) \|_2 \\
    \leq & ~ R^2 \cdot 5 \cdot R_{\infty}^3 \| b \|_2 \sqrt{n} \cdot 2 \sqrt{n} R \exp(R^2) \cdot \| x - y \|_2 \\
    \leq & ~ 80 n^{4} R^6 \exp(4 R^2) \cdot \| x - y \|_2 \\
    \leq & ~ n^{4} \exp(20 R^2) \cdot \| x - y \|_2,
\end{align*}
where the first step is due to the definition of $G_i$ (see Lemma~\ref{lem:summary_G}) and Fact~\ref{fac:basic_matrix_norm}, the second step follows from $\| A \| \leq R$, the third step follows from Lemma~\ref{lem:summary_G}, the fourth step is because of Lemma~\ref{lem:lipschitz_u}, the fifth step is due to the assumption on $R_{\infty}$, and the last step is from simple algebra. 
\end{proof}

\subsection{Lipschitz for \texorpdfstring{$u(x)$}{}}
\label{sub:lipschitz:ux}

We use a tool from \cite{dls23}.
\begin{lemma}[Part 1 of Lemma 7.2 in \cite{dls23}]\label{lem:lipschitz_u}
If the following conditions hold
\begin{itemize}
    \item Let $u(x)$ be defined in definition~\ref{def:u}.
    \item Let $\| A(x - y ) \|_{\infty} < 0.01$
    \item Let $\| A \| \leq R$, where $A \in \R^{n \times d}$
    \item Let $\| x \|_2, \| y \|_2 \leq R$ , where $x, y \in \R^d$
\end{itemize}
then, we have
\begin{align*}
    \| u(x) - u(y) \|_2 \leq 2 \sqrt{n} R \exp(R^2) \| x - y \|_2
\end{align*}
\end{lemma}

\subsection{Lipschitz for \texorpdfstring{$\alpha(x)$}{}}
\label{sub:lipschitz:alphax}

We use a tool from previous work, namely \cite{dls23}.
\begin{lemma}[Part 2 of Lemma 7.2 in \cite{dls23}]\label{lem:lipschitz_alpha}
If the following conditions hold
\begin{itemize}
    \item Let $\alpha(x)$ be defined as Definition~\ref{def:alpha}.
    \item Let $u(x)$ be defined as Definition~\ref{def:u}.
\end{itemize}
then, we have
\begin{align*}
    | \alpha(x) - \alpha(y) | \leq \sqrt{n} \cdot \| u(x) - u(y) \|_2
\end{align*}
\end{lemma}

\subsection{Lipschitz for \texorpdfstring{$c(x)$}{}}
\label{sub:lipschitz:cx}

We find the upper bound of $\|c(x) - c(y)\|_2$.

\begin{lemma}\label{lem:lipschitz_c}
If the following situations hold
\begin{itemize}
    \item Let $c(x)$ be defined in Definition~\ref{def:c}.
    \item Let $\alpha(x)$ be defined as Definition~\ref{def:alpha}.
    \item Let $u(x)$ be defined as Definition~\ref{def:u}.
    \item Let $b \in \R^n$.
\end{itemize}
Then, we have
\begin{align*}
    \| c(x) - c(y) \|_2 \leq \| u (x) - u(y) \|_2 + | \alpha(x) -  \alpha(y) | \cdot \| b \|_2
\end{align*}
\end{lemma}
\begin{proof}

We have
\begin{align*}
\| c(x) - c(y) \|_2 
= & ~ \| ( u(x) - \alpha(x) \cdot b ) - (u(y) - \alpha(y) \cdot b ) \|_2 \\
\leq & ~ \| u(x) - u (y) \|_2 + \| ( \alpha(x) - \alpha(y) ) \cdot b \|_2 \\
= & ~ \| u (x) - u(y) \|_2 + | \alpha(x) -  \alpha(y) | \cdot \| b \|_2
\end{align*}
where the first step is from how we defined $c$ (Definition~\ref{def:c}), the second step is due to the triangle inequality, and the last step follows from simple algebra.
\end{proof}

\subsection{Summary of Five Steps}
\label{sub:lipschitz:summary}

In this section, we analyze the upper bound of the sum of $\|G_i\|$, for all $i \in [5]$.

\begin{lemma}\label{lem:summary_G}
If the following conditions hold
\begin{itemize}
    \item $G_1 = v(x) (v(x) \circ b)^\top - v(y) (v(y) \circ b)^\top$
    \item $G_2 = (v(x) \circ b) v(x)^\top - (v(y) \circ b) v(y)^\top  $ 
    \item $G_3 = \| b \|_2^2 v(x) v(x)^\top - \| b \|_2^2 v(y) v(y)^\top $
    \item $G_4 = \diag((u(x) + c(x)) \circ u(x) ) - \diag((u(y) + c(y)) \circ u(y) )  $ 
    \item $G_5 = \langle c(x), b \rangle \diag(u(x)) - \langle c(y), b \rangle \diag(u(y))$
    \item Let $R_{\infty}:= \max \{ \| u(x) \|_2, \| u (y) \|_2, \| v(x) \|_2, \| v(y)\|_2, \| c(x) \|_2, \| c(y) \|_2, \| b \|_2, 1 \}$
\end{itemize}
Then, we have 
\begin{itemize}
\item Part 1. 
\begin{align*}
 \sum_{i=1}^5 \| G_i \| \leq 20 R_{\infty}^3 \cdot \max\{ \| u(x) - u(y) \|_2 , \| c(x) - c(y) \|_2 \}.
\end{align*}
\item Part 2. Let $\| b \|_2 \leq R$
\begin{align*}
 \sum_{i=1}^5 \| G_i \| \leq 100 R_{\infty}^3 R \sqrt{n} \| u (x)- u(y) \|_2
\end{align*}
\end{itemize}
\end{lemma}
\begin{proof}
{\bf Proof of Part 1.}

Using Lemma~\ref{lem:lipschitz_G_1}, Lemma~\ref{lem:lipschitz_G_2}, Lemma~\ref{lem:lipschitz_G_3}, Lemma~\ref{lem:lipschitz_G_4} and Lemma~\ref{lem:lipschitz_G_5}, we can show for each $i \in [5]$, we have
\begin{align*}
\| G_i \| \leq 20 R_{\infty}^3 \cdot \max\{ \| u(x) - u(y) \|_2, \| c(x) - c(y) \|_2 \}.
\end{align*}

{\bf Proof of Part 2.}

Note that
\begin{align*}
\| c(x) - c(y) \|_2
\leq & ~  \|u(x) - u(y) \|_2 + |\alpha(x) - \alpha(y)| \cdot \| b \|_2 \\
\leq & ~  \|u(x) - u(y) \|_2 + \| u(x) - u(y) \|_2 \sqrt{n} \| b \|_2 \\
\leq & ~  \|u(x) - u(y) \|_2 + \| u(x) - u(y) \|_2 \sqrt{n} R \\
\leq & ~ 2 \sqrt{n} R \| u (x) - u(y) \|_2,
\end{align*}
where the first step follows from Lemma~\ref{lem:lipschitz_c}, the second step follows from Lemma~\ref{lem:lipschitz_alpha}, the third step follows from the assumption on $\| b \|_2 \leq R$, 
and the last step follows from simple algebra.

\end{proof}

\subsection{Lipschitz Calculations: Step 1. Lipschitz for Matrix Function \texorpdfstring{$v(x) (v(x) \circ b)^\top$}{}}
\label{sub:lipschitz:step1}

We find the upper bound of $\|G_1\|$.

\begin{lemma}\label{lem:lipschitz_G_1}
If the following conditions hold
\begin{itemize}
    \item $G_1 = v(x) (v(x) \circ b)^\top - v(y) (v(y) \circ b)^\top$
\end{itemize}
Then, we have
\begin{align*}
    \| G_1 \| \leq 2 \max\{ \| v(x) \|_2, \| v(y) \|_2 \} \cdot \| b \|_2 \cdot \| v( x ) - v ( y ) \|_2.
\end{align*}
\end{lemma}
\begin{proof}
We define
\begin{align*}
G_{1,1} := & ~ v(x) ( v(x) \circ b)^\top - v(y) (v(x) \circ b)^\top \\
G_{1,2} := & ~ v(y) ( v(x) \circ b)^\top - v(y) (v(y) \circ b)^\top
\end{align*}
We have
\begin{align*}
G_1 = G_{1,1} + G_{1,2}
\end{align*}
We can show
\begin{align*}
\| G_{1,1} \| = & ~ \| (v(x) - v(y) ) \cdot (v(x) \circ b)^\top \| \\
\leq & ~ \| v(x) - v(y) \|_2 \cdot \| v(x) \circ b \|_2 \\
\leq & ~ \| v(x) - v(y) \|_2 \cdot \| v (x) \|_2 \cdot \| b \|_2
\end{align*}
where the first step is due to the definition of $G_{1,1}$, the second step follows from Fact~\ref{fac:basic_matrix_norm}, and the last step follows from Fact~\ref{fac:basic_vector_norm}.

Similarly, we can also show
\begin{align*}
\| G_{1,2} \| = & ~ \| v(y) \cdot ( ( v(x) - v(y) ) \circ b  )^\top \| \\
\leq & ~ \| v(y) \|_2 \cdot \| ( v(x) - v(y) ) \circ b \|_2 \\
\leq & ~ \| v(y) \|_2 \cdot \| v(x) - v(y) \|_2 \cdot \| b \|_2
\end{align*}
where the first step is due to the definition of $G_{1,2}$, the second step follows from Fact~\ref{fac:basic_matrix_norm}, and the last step follows from Fact~\ref{fac:basic_vector_norm}.
Thus, we complete the proof.
\end{proof}

\subsection{Lipschitz Calculations: Step 2. Lipschitz for Matrix Function \texorpdfstring{$(v(x) \circ b) v(x)^\top$}{}}
\label{sub:lipschitz:step2}

We look for the upper bound of $\|G_2\|$.

\begin{lemma}\label{lem:lipschitz_G_2}
If the following conditions hold
\begin{itemize}
    \item $G_2 = ( v(x) \circ b) (v(x))^\top - ( v(y) \circ b) v(y)^\top$
\end{itemize}
Then, we have
\begin{align*}
    \| G_2 \| \leq 2 \max\{ \| v(x) \|_2, \| v(y) \|_2 \} \cdot \| b \|_2 \cdot \| v( x ) - v( y ) \|_2.
\end{align*}
\end{lemma}
\begin{proof}
The proof is very similar to the previous Lemma. So we omit the details here.
\end{proof}

\subsection{Lipschitz Calculations: Step 3. Lipschitz for Matrix Function \texorpdfstring{$\| b \|_2^2 v(x)  v(x)^\top$}{}}
\label{sub:lipschitz:step3}

We analyze the upper bound of $\|G_3\|$.

\begin{lemma}\label{lem:lipschitz_G_3}
If the following conditions hold
\begin{itemize}
    \item $G_3 = \| b \|_2^2 v(x)  v(x)^\top - \| b \|_2^2 v(y)  v(y)^\top$
\end{itemize}
Then, we have
\begin{align*}
    \| G_3 \| \leq 2 \max\{ \| v(x) \|_2, \| v(y) \|_2 \} \cdot \| b \|_2^2 \cdot \| v( x ) - v( y ) \|_2.
\end{align*}
\end{lemma}
\begin{proof}
We define
\begin{align*}
G_{3,1} := & ~ \| b \|_2^2 v(x) v(x)^\top - \| b \|_2^2 v(y) v(x)^\top \\
G_{3,2} := & ~ \| b \|_2^2 v(y) v(x)^\top - \| b \|_2^2 v(y) v(y)^\top
\end{align*}
We have
\begin{align*}
G_3 = G_{3,1} + G_{3,2}.
\end{align*}

We can show that
\begin{align*}
\| G_{3,1} \| 
= & ~ \| b \|_2^2 \cdot \| v(x) v(x)^\top - v(y) v(x)^\top \| \\
= & ~ \| b \|_2^2 \cdot \| (v(x) - v(y)) v(x)^\top \| \\
\leq & ~ \| b \|_2^2 \cdot \| v(x) - v (y) \|_2 \cdot \| v (x) \|_2,
\end{align*}
where the first step comes from the definition of $G_{3,1}$, the second step is due to simple algebra, and the third step follows from Fact~\ref{fac:basic_matrix_norm}.

Similarly, we can show that
\begin{align*}
    \| G_{3,2} \| \leq \| b \|_2^2 \cdot \| v(x) - v (y) \|_2 \cdot \| v (x) \|_2.
\end{align*}
Thus, we complete the proof.
\end{proof}

\subsection{Lipschitz Calculations: Step 4. Lipschitz for Matrix Function \texorpdfstring{$\diag( (u(x) + c(x)) \circ u(x) )$}{}}
\label{sub:lipschitz:step4}

We show the upper bound of $\|G_4\|$.

Since we need to prove the Lipschitz, the effect of $q$ make no difference. The $q$ will be canceled. Thus, we define the terms without having $q$ at all.
\begin{lemma}\label{lem:lipschitz_G_4}
If the following conditions hold
\begin{itemize}
    \item $G_4 = \diag( ( u(x) +c(x) ) \circ u(x)  ) - \diag( ( u(y) +c(y) ) \circ u(y)  ) $
\end{itemize}
Then, we have
\begin{align*}
     \| G_4 \| \leq 4 \max\{ \| u(x) \|_2, \| u(y) \|_2, \| c(x) \|_2, \| c(y) \|_2 \} \cdot ( \| u (x) - u (y) \|_2 + \| c(x) - c(y) \|_2 )
\end{align*}
\end{lemma}
\begin{proof}
We define
\begin{align*}
G_{4,1} := & ~ \diag( ( u(x) + c(x) ) \circ u(x)  ) - \diag( (u(y) + c(y)) \circ u(x) ) \\
G_{4,2} := & \diag( (u(y) + c(y)) \circ u(x) ) - \diag( (u(y) + c(y)) \circ u(y) )
\end{align*}

Then we have 
\begin{align*}
\| G_{4,1} \| = & ~ \| \diag( ( u(x) + c(x) ) \circ u(x)  ) - \diag( (u(y) + c(y)) \circ u(x) ) \| \\
\leq & ~ \|  ( u(x)+c(x) - u(y) - c(y) ) \circ u(x) \|_{2} \\
\leq & ~  \|   u(x)+c(x) - u(y) - c(y) \|_2 \cdot \| u(x) \|_2 \\
\leq & ~ ( \| u (x) - u (y) \|_2 + \| c(x) - c(y) \|_2 ) \cdot \| u (y) \|_2
\end{align*}
where the first step is due to the  definition of $G_{4,1}$, the second step is due to Fact~\ref{fac:basic_vector_norm}, and the third step is due to Fact~\ref{fac:basic_vector_norm} and the last step follows from triangle inequality.

Similarly, we have
\begin{align*}
\| G_{4,2} \| = & ~ \| \diag( (u(y) + c(y)) \circ u(x) ) - \diag( (u(y) + c(y)) \circ u(y) ) \| \\
\leq & ~ \|  (u(y) + c(y)) \circ u(x) - (u(y) + c(y)) \circ u(y)  \|_2 \\
\leq & ~ (\| u(y) \|_2 + \| c(y) \|_2 ) \cdot \| u(x) - u(y) \|_2
\end{align*}
where the first step is due to the  definition of $G_{4,2}$, the second step is due to Fact~\ref{fac:basic_vector_norm}, and the third step is due to Fact~\ref{fac:basic_vector_norm}.
\end{proof}

\subsection{Lipschitz Calculations: Step 5. Lipschitz for Matrix Function \texorpdfstring{$\langle c(x), b \rangle \diag(u(x))$}{}}
\label{sub:lipschitz:step5}

We compute the upper bound of $\|G_5\|$.

\begin{lemma}\label{lem:lipschitz_G_5}
If the following conditions hold
\begin{itemize}
    \item $G_5 = \langle c(x), b \rangle \diag(u(x)) - \langle c(y), b \rangle \diag(u(y))$
\end{itemize}
Then, we have
\begin{align*}
     \| G_5 \| \leq 4 \max\{ \| u(x) \|_2, \| u(y) \|_2, \| c(x) \|_2, \| c(y) \|_2 \} \cdot \| b \|_2 ( \| u (x) - u (y) \|_2 + \| c(x) - c(y) \|_2 )
\end{align*}
\end{lemma}
\begin{proof}
We define
\begin{align*}
G_{5,1} := & ~ \langle c(x), b \rangle \diag(u(x)) - \langle c(x), b \rangle \diag(u(y)) \\
G_{5,2} := & ~ \langle c(x), b \rangle \diag(u(y)) - \langle c(y), b \rangle \diag(u(y)) \\
\end{align*}
We can show
\begin{align*}
\| G_{5,1} \| 
= & ~ \| \langle c(x), b \rangle \cdot ( \diag(u(x)) - \diag(u(y)) ) \| \\
= & ~  | \langle c(x), b \rangle | \cdot \| \diag(u(x)) - \diag(u(y)) \| \\
\leq & ~ \| c(x) \|_2 \cdot \| b \|_2 \cdot \| \diag(u(x)) - \diag(u(y)) \| \\
\leq & ~ \| c(x) \|_2 \cdot \| b \|_2 \cdot \| u(x) - u(y) \|_2
\end{align*}
where the first step is due to the definition of $G_{5,1}$, the second step follows from Fact~\ref{fac:basic_matrix_norm}, the second step follows from Fact~\ref{fac:basic_vector_norm}, and the last step follows from Fact~\ref{fac:basic_vector_norm}.

Similarly, we have
\begin{align*}
\| G_{5,2} \| = & ~ | \langle c(x) - c(y), b \rangle | \cdot \| \diag(u(y) ) \| \\
\leq & ~ \| c(x) - c(y) \|_2 \cdot \| b \|_2 \cdot \| u(y) \|_2
\end{align*}
where the first step is due to Fact~\ref{fac:basic_vector_norm}, the definition of $G_{5,2}$ and simple algebra, and the second follows from Fact~\ref{fac:basic_vector_norm} and Fact~\ref{fac:psd}. 
\end{proof}

%% file: lip_A.tex
\section{Lipschitz with Respect To \texorpdfstring{$A$}{}}\label{sec:lipschitz:A}

In Section~\ref{sub:lipschitzA:x}, we consider the $x$ case, which is to upper bound $|\alpha(x)^{-1}|$.
In Section~\ref{sub:lipschitzA:A}, we consider the $A$ case, namely computing the upper bound of $|\alpha(A)^{-1}|$.
In Section~\ref{sub:lipschitzA:uA}, we analyze the bound for $\|u(A) - u(B)\|_2$.
In Section~\ref{sub:lipschitzA:alphaA}, we prove the bound for $|\alpha(A) - \alpha(B)|$.
In Section~\ref{sub:lipschitzA:cx}, we analyze the bound for $\|c(A) - c(B)\|_2$.

\subsection{For the \texorpdfstring{$x$}{} case}
\label{sub:lipschitzA:x}

In this section, the goal is to bound $|\alpha(x)^{-1}|$. We start from the following definition.

\begin{definition}\label{def:delta_b}
We define $\delta_b$ be to the vector that satisfies
\begin{align*}
\| u(x_{t+1}) - \alpha(x_{t+1}) b \|_2^2 = \| u(x_t) - \alpha(x_t) (b - \delta_b) \|_2^2
\end{align*}
\end{definition}

\begin{lemma}\label{lem:rewrite_delta_b}
We have 
\begin{align*}
\| \delta_b \|_2 \leq |\alpha(x_t)^{-1}| \cdot \| c(x_{t+1} ) - c(x_t) \|_2
\end{align*}
\end{lemma}
\begin{proof}
Similarly as \cite{lsz+23} described, there could be multiple solutions, e.g. $2^n$ possible solutions
\begin{align*}
u(x_{t+1}) - \alpha(x_{t+1}) b =  ( u(x_t) - \alpha(x_t) (b - \delta_b) ) \circ \{-1,+1\}^n
\end{align*}
The norm of all the solutions are same. Therefore, we can just consider one solution, which is the following solution 
\begin{align*}
u(x_{t+1}) - \alpha(x_{t+1}) b = u(x_t) - \alpha(x_t) (b - \delta_b)
\end{align*}
Thus, 
\begin{align*}
\delta_b = & ~ \alpha(x_t)^{-1} ( u(x_{t+1}) - u(x_t) - b ( \alpha(x_{t+1}) - \alpha(x_t) ) ) \\
= & ~ \alpha(x_t)^{-1} (c (x_{t+1}) - c(x_t) )
\end{align*}
\end{proof}

We use a tool, which is from \cite{dls23}.
\begin{lemma}[Lemma 8.9 in \cite{dls23}]\label{lem:bound_alpha_inverse}
If the following condition hold
\begin{itemize}
    \item Let $\| A \| \leq R$
    \item Let $\| x \|_2 \leq R$
\end{itemize}
We have
\begin{align*}
    | \alpha(x)^{-1} | \leq \exp(R^2)
\end{align*}
\end{lemma}
The proof is standard, we omit the details here.

\subsection{For the \texorpdfstring{$A$}{} case}
\label{sub:lipschitzA:A}

Here, we bound $| \alpha(A)^{-1} |$.

\begin{definition}\label{def:delta_b:A}
We define $\delta_b$ be to the vector that satsifies
\begin{align*}
\| u(x_{t+1}) - \alpha(x_{t+1}) b \|_2^2 = \| u(x_t) - \alpha(x_t) (b - \delta_b) \|_2^2
\end{align*}
\end{definition}

\begin{lemma}\label{lem:rewrite_delta_b:A}
We have 
\begin{align*}
\| \delta_b \|_2 \leq |\alpha(x_t)^{-1}| \cdot \| c(x_{t+1} ) - c(x_t) \|_2
\end{align*}
\end{lemma}

\begin{lemma}[Lemma 8.9 in \cite{dls23}]\label{lem:bound_alpha_inverse:A}
If the following points hold
\begin{itemize}
    \item Let $\| A \| \leq R$
    \item Let $\| x \|_2 \leq R$
\end{itemize}
We have
\begin{align*}
    | \alpha(A)^{-1} | \leq \exp(R^2)
\end{align*}
\end{lemma}

\subsection{Lipschitz for \texorpdfstring{$u(A)$}{}}
\label{sub:lipschitzA:uA}

We state a tool that directly implies by previous work. The proof is very standard, so we omit the details here.

\begin{lemma}[A variation of Part 1 of Lemma 7.2 in \cite{dls23}]\label{lem:lipschitz_u:A}
If the following conditions hold
\begin{itemize}
    \item Let $u(A)$ be defined as definition~\ref{def:u} with reparamerization by $A$ instead of $x$.\footnote{Instead of calling $u(x) = \exp(Ax)$. We call $u(A) = \exp(Ax)$.} 
    \item Let $\| (A-B) x  \|_{\infty} < 0.01$
    \item Let $\| A \| , \| B \| \leq R$, where $A, B \in \R^{n \times d}$
    \item Let $\| x \|_2\leq R$ , where $x\in \R^d$
\end{itemize}
then, we have
\begin{align*}
    \| u(A) - u(B) \|_2 \leq 2 \sqrt{n} R \exp(R^2) \| A - B \|
\end{align*}
\end{lemma}

\subsection{Lipschitz for \texorpdfstring{$\alpha(A)$}{}}

\label{sub:lipschitzA:alphaA}

We state a tool which directly implies by previous work. The proof is very standard, so we omit the details here.
\begin{lemma}[A variation of Part 2 of Lemma 7.2 in \cite{dls23}]\label{lem:lipschitz_alpha:A}
If the following conditions hold
\begin{itemize}
    \item Let $\alpha(A)$ be defined in Definition~\ref{def:alpha} with reparameterization by $A$ instead of $x$.
    \item Let $u(A)$ be defined as Definition~\ref{def:u} with reparameterization by $A$ instead of $x$.
\end{itemize}
then, we have
\begin{align*}
    | \alpha(A) - \alpha(B) | \leq \sqrt{n} \cdot \| u(A) - u(B) \|_2
\end{align*}
\end{lemma}

\subsection{Lipschitz for \texorpdfstring{$c(x)$}{}}
\label{sub:lipschitzA:cx}

In this section, we bound $\| c(A) - c(B) \|_2$.

\begin{lemma}[A variation of Lemma~\ref{lem:lipschitz_c}]\label{lem:lipschitz_c:A}
If the following conditions hold
\begin{itemize}
    \item Let $c(A)$ be defined as Definition~\ref{def:c} with reparametrization by $A$.
    \item Let $\alpha(A)$ be defined as Definition~\ref{def:alpha} with reparameterization by $A$.
    \item Let $u(A)$ be defined as Definition~\ref{def:u} with reparameterization by $A$.
    \item Let $b \in \R^n$.
\end{itemize}
Then, we have
\begin{align*}
    \| c(A) - c(B) \|_2 \leq \| u (A) - u(B) \|_2 + | \alpha(A) -  \alpha(B) | \cdot \| b \|_2
\end{align*}
\end{lemma}
\begin{proof}

We have
\begin{align*}
\| c(A) - c(B) \|_2 
= & ~ \| ( u(A) - \alpha(B) \cdot b ) - (u(A) - \alpha(B) \cdot b ) \|_2 \\
\leq & ~ \| u(A) - u (B) \|_2 + \| ( \alpha(A) - \alpha(B) ) \cdot b \|_2 \\
= & ~ \| u (A) - u(B) \|_2 + | \alpha(A) -  \alpha(B) | \cdot \| b \|_2
\end{align*}
where the first step comes from how we defined $c$ (see Definition~\ref{def:c}), the second step is because of the triangle inequality, and the last step follows from simple algebra.
\end{proof}

%% file: result.tex
\section{Main Result}\label{sec:main_result}

In Section~\ref{sub:main_result:convergence}, we introduce our algorithm (see Algorithm~\ref{alg:main:informal}) and use our main Theorem (see Theorem~\ref{thm:main_formal}) to analyze its properties, including running time and the output $\wt{x}$.
In Section~\ref{sub:main_result:application}, we introduce a corollary which is the application of in-context learning.

\subsection{Convergence}
\label{sub:main_result:convergence}

Now, we introduce our main algorithm and Theorem.

\begin{theorem}[ 
]\label{thm:main_formal}
Given that vectors $b,w \in \R^n$ and a matrix $A \in \R^{n \times d}$, we define $x^*$ as the optimal solution of the following problem
\begin{align*}
     \min_{x \in \R^d} 0.5 \cdot \|  \exp(A x) - \langle \exp(Ax) , {\bf 1}_n \rangle \cdot b \|_2^2+0.5 \| \diag(w) A x\|_2^2  
\end{align*}
And then if the conditions as follows hold:
\begin{itemize}
    \item $R \geq 4$. 
    \item $g(x^*) = {\bf 0}_d$.
    \item $\| x^* \|_2 \leq R$.
    \item $\| A \| \leq R$.
    \item $\| b \|_2  \leq R$.
    \item $w_{i}^2 \geq 100 + l/\sigma_{\min}(A)^2$ for all $i \in [n]$
    \item $M = \exp(O( R^2 + \log n))$.
    \item Let accuracy $\epsilon \in (0,0.1)$
    \item Let failure probability $\delta \in (0,0.1)$
    \item Let $x_0$ denote an initial point for which it holds that $M \| x_0 - x^* \|_2 \leq 0.1 l$.
\end{itemize}
Then there exists a randomized algorithm (Algorithm~\ref{alg:main:informal}) such that, with probability at least $1-\delta$,
\begin{itemize}
    \item it runs $T = \log(\| x_0 - x^* \|_2/ \epsilon)$ iterations
    \item spends 
    \begin{align*}
            O( (\nnz(A) + d^{\omega} ) \cdot \poly(\log(n/\delta)). 
    \end{align*}
    \item outputs a vector $\wt{x} \in \R^d$ such that 
    \begin{align*}
        \| \wt{x} - x^* \|_2 \leq \epsilon
    \end{align*}
\end{itemize}
Here $\omega$ denote the exponent of matrix multiplication. Currently $\omega \approx 2.373$ \citep{w12,lg14,aw21}.  
 
\end{theorem}

\begin{proof}

{\bf Proof of Hessian is PD.}  

We can obtain this conclusion from  Lemma~\ref{lem:hessian_psd_exp}.  

{\bf Proof of Hessian is Lipschitz.}

The proof is due to  Lemma~\ref{lem:hessian_lipschitz}.

{\bf Proof of Cost per iteration.}

This follows from Lemma~\ref{lem:subsample}.

{\bf Proof of Convergence per Iteration.}

By Lemma~\ref{lem:one_step_shrinking}, we have
\begin{align*}
    \|x_k - x^*\|_2 \le 0.4 \cdot \|x_{k-1} - x^*\|_2.
\end{align*}

{\bf Proof of Number of Iterations.}

    After $T$ iterations, we have
    \begin{align*}
    \| x_T - x^* \|_2 \leq 0.4^T \cdot \| x_0 - x^* \|_2
    \end{align*}
    By choice of $T$, we get the desired bound. The failure probability is following from union bound over $T$ iterations.

\end{proof}

\subsection{Application to In-context Learning}
\label{sub:main_result:application}

In this section, we introduce the application to in-context learning.

\begin{corollary}[Bounded shift for Learning in-context]\label{cor:in_context_learning}
If the following conditions hold
\begin{itemize}
    \item Let $A \in \R^{n \times d}$.
    \item Let $b \in \R^n$.
    \item $\| A \| \leq R$.
    \item Let $\| x \|_2 \leq R$.
    \item $\| A (x_{t+1} - x_t) \|_{\infty} < 0.01$.
    \item $\| (A_{t+1} - A_t) x \|_{\infty} < 0.01$.
    \item Let $R \geq 4$.
    \item Let $M:= \exp(O( R^2 + \log n))$.
    \item Let $u(x) \in \{\exp(Ax), \cosh(Ax), \sinh(Ax) \}$.
\end{itemize}
We consider the rescaled softmax regression (Definition~\ref{def:rescaled_softmax_regression}) problem
\begin{align*}
    \min_{x \in \R^d} \| u(x) - \alpha (x) b \|_2.
\end{align*}
\begin{itemize}
\item {\bf Part 1.} If we move the $x_t$ to $x_{t+1}$, then we're solving a new rescaled softmax regression problem with
\begin{align*}
    \min_{x \in \R^d } \| u(x) - \alpha(x) \wt{b} \|_2
\end{align*}
where 
\begin{align*}
    \| \wt{b} - b \|_2 \leq M \cdot \| x_{t+1} - x_t \|_2
\end{align*}
\item {\bf Part 2.} If we move the $A_t$ to $A_{t+1}$, then we're solving a new rescaled softmax regression with
\begin{align*}
     \min_x \| u(x) - \alpha(x) \wh{b} \|_2
\end{align*}
where 
\begin{align*}
    \| \wh{b} - b \|_2 \leq M \cdot \| A_{t+1} - A_t \|
\end{align*}
\end{itemize}
\end{corollary}
\begin{proof}
{\bf Proof of Part 1.}
The proof follows from by combining Lemma~\ref{lem:rewrite_delta_b}, Lemma~\ref{lem:bound_alpha_inverse}, Lemma~\ref{lem:lipschitz_u}, Lemma~\ref{lem:lipschitz_alpha}, Lemma~\ref{lem:lipschitz_c}.

{\bf Proof of Part 2. }
The proof follows from by combining Lemma~\ref{lem:rewrite_delta_b:A}, Lemma~\ref{lem:bound_alpha_inverse:A}, Lemma~\ref{lem:lipschitz_u:A}, Lemma~\ref{lem:lipschitz_alpha:A}, Lemma~\ref{lem:lipschitz_c:A}.
\end{proof}

\section{More Related Works}
\label{sub:related_work:iteration}

One of the important ideas in this work is to use sketching to speed up the iterative algorithm in optimization. The fundamental concept of sketching is to decompose a large input matrix into a significantly smaller sketching matrix, but this sketching matrix retains the important characteristics of the original matrix. Therefore, the algorithms can only operate on this smaller matrix instead of the unwieldy original one, resulting in a substantial reduction in computational time. There are numerous prior studies that have devised sketching algorithms with robust theoretical assurances. For example, the Johnson-Lindenstrauss lemma in \cite{jl84} demonstrates that, in certain high-dimensional spaces, projecting points onto lower-dimensional subspaces can preserve pairwise distances between the points. This property supports the development of faster algorithms for tasks such as nearest neighbor search. Furthermore, as elucidated in \cite{ac06}, the Fast Johnson-Lindenstrauss Transform (FJLT) introduces a specific family of structured random projections that can be applied to an input matrix in time proportional to its sparsity. 

There are two ways to utilize the sketching matrices. The first way is known as sketch-and-solve, which uses sketching a predetermined number of times. This may lead to faster algorithms in several domains, like in the linear regression \citep{nn13,cw13} and low-rank approximation \citep{syyz23_weighted}, in column subset selection \citep{sg22,swz19b,jll+20,jll+21}, where, with provable approximation guarantees, the column selection can be speed up by sketching the data matrix, in kernel methods \citep{lgtc15}, where the sketching methods can be applied to large kernel matrices approximation, in tensor method \citep{anw14,p13,pp13,dssw18,dsy23}, tensors can be compressed down to much smaller core tensors. Additionally, it can be employed to determine the optimal bound as demonstrated in \cite{syyz23_linf} and to design an efficient method for training neural networks, as shown in \cite{qsy23}. Moreover, a recent work \citep{syz23_quantum} has applied the sketching method to the quantum algorithm, which solves the linear regression problem. Finally, it has been used to study the matrix completion problem in \cite{gsyz23}.

The second way is known as iterate-and-sketch, which is applied in each iteration of the optimization algorithm and establishes a robust analysis framework. It has been widely used in numerous important tasks such as linear programming \citep{jswz21,sy21,lsz+23,cls19,gs22}, empirical risk minimization \citep{lsz19,qszz23}, John Ellipsoid algorithm \citep{syyz22}, online weighted matching problem \citep{swyy23}, the Frank-Wolfe algorithm \citep{sxyz22,xss21}, semidefinite programming \citep{gs22,syyz23}, federated learning \citep{swyz23,bsy23}, attention approximation \citep{gsy23_coin,gswy23}, $k$ means clustering \citep{lss+22}, discrepancy algorithm \citep{dsw22}, training over-parametrized neural network \citep{szz21,als+22,z22}, rational database \citep{qjs+22}, matrix sensing \citep{qsz23}.

Other theoretical machine learning works focus on LLMs efficiency \citep{zyw+25,chen2025universal,chen2024fast,xsw+24,xsl24,llss24_softmax,lssz24_tat,swxl24,lssy24,wms+24,lssz24_dp,llss24_sparse,lss+24,smn+24,lls+24_io,kll+24,cll+24_rope,chl+24_rope,lls+24_tensor,lls+25_prune,cll+25_icl,lls+25_grok,kls+25,cls+25,lll+25_loop,cll+25_var,ccl+25,chl+24_gat,zlz21}, reinforcement learning \citep{reinforcement_1,reinforcement_2,reinforcement_3,reinforcement_4,reinforcement_5,reinforcement_6}, circuit complexity \citep{circuit1,lls+25_graph}, fairness analysis \citep{cll+25_fair}, and differential privacy \citep{dp1,dp2,dp3,dp4,dp5,dp6,dp7}.